\documentclass[11pt]{article}

\usepackage[final]{acl}

\usepackage{times}
\usepackage{latexsym}
\usepackage[T1]{fontenc}

\usepackage[utf8]{inputenc}

\usepackage{microtype}

\usepackage{inconsolata}

\usepackage{graphicx}
\usepackage{booktabs}  
\usepackage{tabularx}
\usepackage{makecell}
\usepackage{array}
\usepackage{amsmath,amssymb} 
\usepackage{float}

\usepackage{subcaption}
\usepackage{caption}

\usepackage[most]{tcolorbox} 
\tcbset{
  aibox/.style={
    width=\linewidth,
    top=8pt,
    bottom=4pt,
    colback=blue!5!white,
    colframe=black!85,
    colbacktitle=black, 
    enhanced,
    center,
    attach boxed title to top left={yshift=-0.1in,xshift=0.15in},
    boxed title style={boxrule=0pt,colframe=white,},
  }
}
\newtcolorbox{AIbox}[2][]{aibox,title=#2,#1}

\title{A Syllogistic Probe: Tracing the Evolution of Logic Reasoning in Large Language Models}

\author{
 \textbf{Zhengqing Zang\textsuperscript{1,3*}},
 \textbf{Yuqi Ding\textsuperscript{2,3*}},
 \textbf{Yanmei Gu\textsuperscript{3$\dagger$}},
 \textbf{Changkai Song\textsuperscript{3}},
 \textbf{Zhengkai Yang\textsuperscript{3}},
 \textbf{Guoping Du\textsuperscript{4}},
 \\
 \textbf{Junbo Zhao\textsuperscript{1,3}},
 \textbf{Haobo Wang\textsuperscript{1$\dagger$}}
 \\
 \textsuperscript{1}Zhejiang University,
 \textsuperscript{2}University of Chinese Academy of Social Sciences,
 \\
 \textsuperscript{3}Ant Group,
 \textsuperscript{4}Chinese Academy of Social Sciences
 \\
 \small{
\{zangzq, wanghaobo\}@zju.edu.cn, dingyuqi@ucass.edu.cn, 
yanmeigu.gym@antgroup.com
 }
}

\begin{document}
\maketitle

\begingroup
\renewcommand{\thefootnote}{}  
\footnotetext{\textsuperscript{*}These authors contributed equally.}
\footnotetext{\textsuperscript{$\dagger$}Corresponding Authors.}
\endgroup
\setcounter{footnote}{0} 

\begin{abstract}

Human logic has gradually shifted from intuition-driven inference to rigorous formal systems. Motivated by recent advances in large language models (LLMs), we explore whether LLMs exhibit a similar evolution in the underlying logical framework. Using existential import as a probe, we for  evaluate syllogism under traditional and modern logic. Through extensive experiments of testing SOTA LLMs on a new syllogism dataset, we have some interesting findings: (i) Model size scaling promotes the shift toward modern logic; (ii) Thinking serves as an efficient accelerator beyond parameter scaling; (iii) the Base model plays a crucial role in determining how easily and stably this shift can emerge. 
Beyond these core factors, we conduct additional experiments 
for in-depth analysis of properties of current LLMs on syllogistic reasoning.

\end{abstract}

\section{Introduction}

Human logic has evolved from earlier, more intuition-driven accounts of valid inference~\cite{Aristotle} to increasingly rigorous formal frameworks~\cite{enderton-math-logic}. In particular, the development of symbolic logic clarified the semantics of quantification and enabled precise validity checking under explicit model-theoretic interpretations, laying the foundation for contemporary logical analysis.

Recently, neural networks have evolved from early, relatively simple architectures with limited capacity for logical reasoning to today’s large language models (LLMs), which have achieved remarkable progress across natural language processing tasks. State-of-the-art models such as GPT-5 \cite{openai_gpt5} and Gemini-3-Pro-Preview \cite{gemini_3_pro}, often rival human experts in complex reasoning tasks ranging from commonsense reasoning \cite{bang2023multitask, bisk2019piqa} to
mathematical problem-solving \cite{phan2025humanitysexam, wei2023cmath}. These advances raise a natural question: \emph{do LLMs exhibit an analogous evolution in their underlying logical framework?} If so, \emph{what changes, and how does this change emerge?}


\begin{figure}[t]
    \centering
    \begin{subfigure}[t]{\linewidth}
        \centering
        \includegraphics[width=\linewidth]{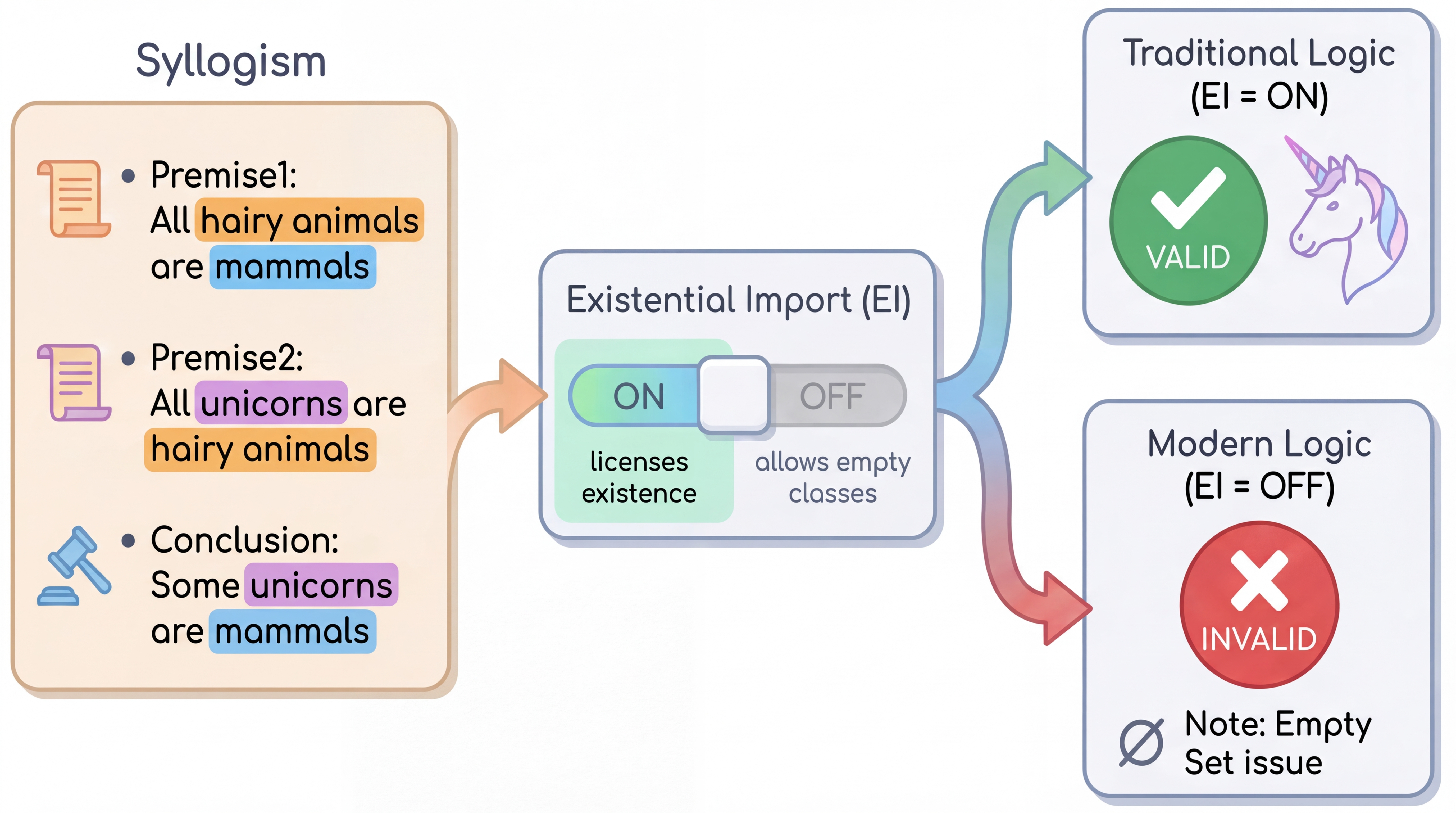}
    \end{subfigure}
    \begin{subfigure}[t]{\linewidth}
        \centering
        \includegraphics[width=\linewidth]{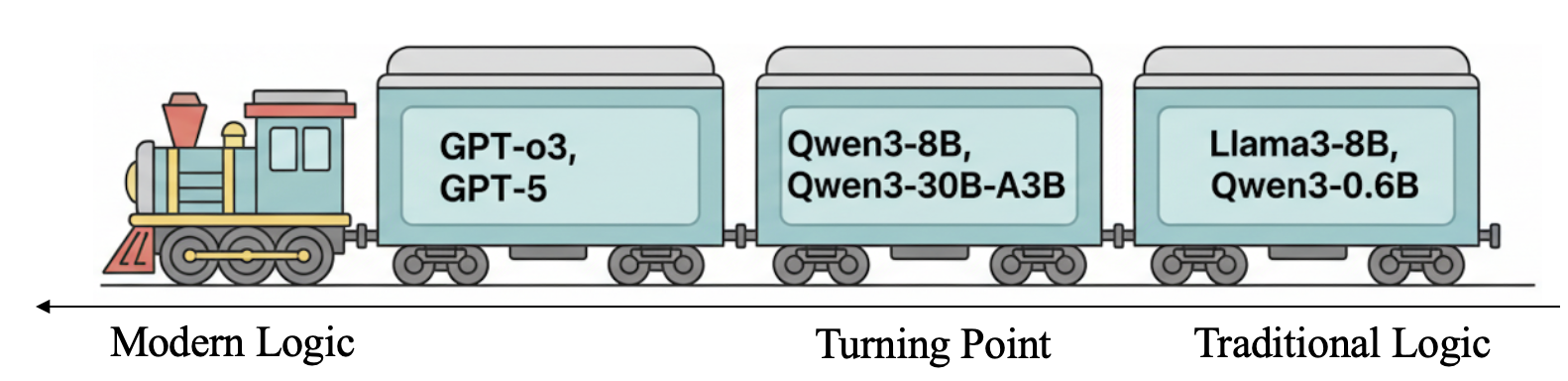}
    \end{subfigure}
    \caption{The illustration of existential import problem and the trace of model logic.
}
\label{fig:existential import problem}
\end{figure}

\begin{figure*}[t]
    \centering
    \begin{subfigure}[t]{0.95\textwidth}
        \centering
        \includegraphics[width=\linewidth]{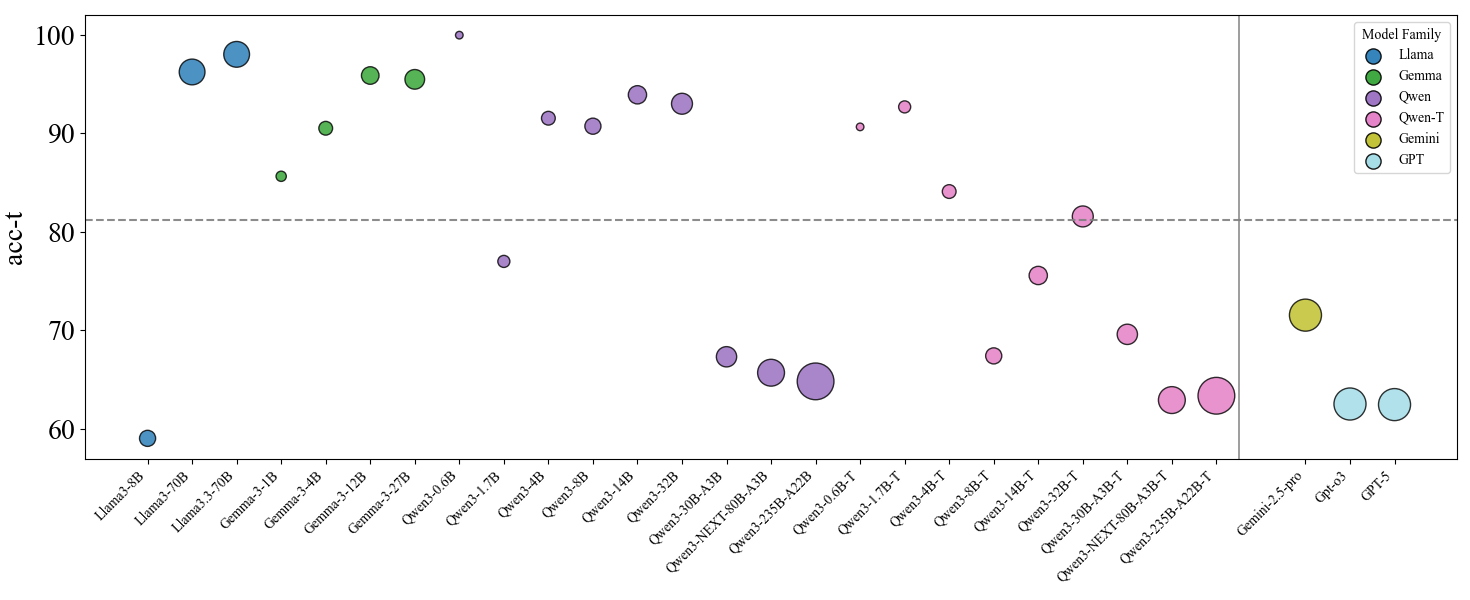}
    \end{subfigure}
    \begin{subfigure}[t]{0.95\textwidth}
        \centering
        \includegraphics[width=\linewidth]{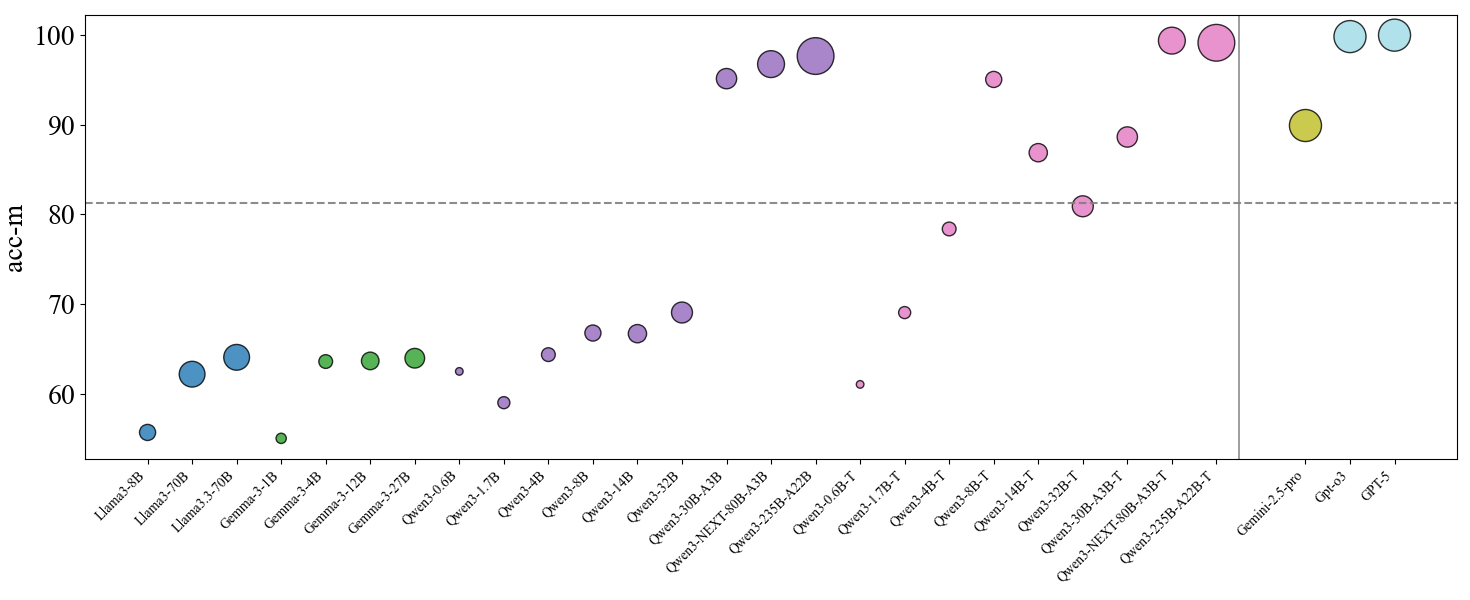}
    \end{subfigure}
    \caption{Overall performance of auto-regressive models under traditional logic and modern logic. The upper figure shows model performance under the traditional logic criterion, while the lower panel reports performance under the modern logic criterion. Point size is proportional to model scale, and color denotes model family. \texttt{Qwen-T} indicates Qwen \texttt{Thinking} models/mode. For closed-source models, we use a fixed medium point size for visualization only, which does not reflect their true parameter counts. The horizontal dashed line marks the dividing line between the traditional and modern logic.
}
    \label{fig:stacked_two}
\end{figure*}

Existing reasoning benchmarks~\cite{han2024folio} increasingly target first-order logic(modern logic), examining whether they can follow this more rigorous, formal style of reasoning. However, in syllogism reasoning, existing datasets~\cite{ando2023NeuBAROCO, nguyen2025bisreasoning, wu-etal-2023-hence} typically treat traditional logic as the implicit default. This creates a systematic bias. A model may score high simply because it has learned dataset-specific shortcuts in traditional syllogisms, not because it truly has rigorous reasoning ability that can be transferred to new settings. On the other hand, a model may score low because it takes a modern logic view and therefore refuses to infer existence from a statement like “All unicorns are hairy animals”, which then gets marked as wrong. Even worse, these unstated rules mix up a model’s reasoning ability with how well it matches the evaluation convention, making the scores hard to interpret.


In this study, we focus on syllogism~\cite{Aristotle}, a classic and well-studied form of deductive reasoning.  The evaluation of syllogisms differs between two frameworks: \textbf{Traditional Logic} (Aristotelian) and \textbf{Modern Logic} (Boolean interpretation). The key difference between them lies in \textbf{existential import (EI)} \cite{sep-square}, where traditional logic typically assumes the relevant terms are non-empty while modern logic makes existential commitments only when explicitly stated. As shown in Figure \ref{fig:existential import problem}, this syllogism is typically treated as "valid" in traditional logic because the universal statement about unicorns \textit{(Premise2: All unicorns are hairy animals)} is taken to presuppose that unicorns exist. While in modern logic, the conclusion does not follow unless existence of unicorns is separately asserted \textit{("Some unicorns exist")}, since the premise can be true even when there are no unicorns. 

To trace the evolution of logic reasoning in LLMs, we use existential import as a probe and conduct a series of investigations on a new syllogism dataset, which can be summarized in following key findings:



\textbf{(1) Controlled evidence across open-source model families and scales.}
We run systematic evaluations on Qwen 3~\cite{yang2025qwen3technicalreport}, Llama 3~\cite{grattafiori2024llama3herdmodels} , and Gemma 3~\cite{gemmateam2025gemma3technicalreport} series across model sizes and training variants. 
We find that as model size increases, $acc_m$ rises across all models. 
Models in Llama 3 and Gemma 3 series largely retain a traditional-logic reasoning style. However, in the Qwen series, we observe a clear shift in its logical paradigm from traditional logic to modern logic. We also identify a turning point where consistency fluctuates during the transition.

\textbf{(2) Thinking as an efficient driver beyond parameter scaling.}
By comparing matched-size models, we show that RL-trained thinking variants can strongly accelerate the shift toward modern logic and improve consistency. Prompted chain-of-thought yields only a partial shift, and distillation alone does not reliably produce strict modern logic behavior, suggesting that the transition is driven more by post-training optimization of reasoning policies than by scale or imitation learning alone.

\textbf{(3) Base-model constraints on learnability and stability.}
We evaluate Base models and show that they set the starting point for post-training. When the Base model already shows signals aligned with modern logic, post-training shifts are easier and more stable. Otherwise, the shift is harder and less stable.

We further report the experiments of different prompts, the emptiness of minor terms, cross-lingual gaps, and architecture effects including diffusion-based LLMs, conducting an in-depth analysis of properties of current LLMs on syllogistic reasoning.

\section{Background and Dataset Construction}

\subsection{Syllogism and Existential Import}
\label{section2}

Aristotle characterizes a syllogism as consisting of two premises and a conclusion \cite{Aristotle}, where each statement is a categorical proposition relating a subject term ($S$) to a predicate term ($P$). Within the syllogism's structure, the conclusion's subject ($S$) is called the minor term, and its predicate ($P$) the major term. In standard form, there are four categorical proposition types (A/E/I/O):
\begin{align*}
\text{A(universal affirmative):} &\ \text{All $S$ are $P$} \\
\text{E(universal negative):} &\ \text{No $S$ are $P$} \\
\text{I(particular affirmative):} &\ \text{Some $S$ are $P$} \\
\text{O(particular negative):} &\ \text{Some $S$ are not $P$}
\end{align*}

In this paper, we use \textbf{traditional logic} to denote the Aristotelian syllogistic framework, and \textbf{modern logic} to denote the Boolean interpretation of categorical propositions \cite{Boole1854}. For reference, under modern logic these four forms are typically rendered as:
\begin{align*}
\text{A:} &\ \forall x\,(Sx \rightarrow Px) \\
\text{E:} &\ \forall x\,(Sx \rightarrow \neg Px)\ \text{(equiv.\ } \neg \exists x\,(Sx \land Px)\text{)} \\
\text{I:} &\ \exists x\,(Sx \land Px) \\
\text{O:} &\ \exists x\,(Sx \land \neg Px)
\end{align*}

The core distinction is \textbf{existential import (EI)}: whether a proposition is taken to imply that its subject class is non-empty \cite{sep-square}.

\noindent$\bullet$ In \textbf{traditional logic}, universal propositions (A/E) are typically assumed to have EI: for instance, "All $S$ are $P$" is read as implying that the class $S$ is not empty.

\noindent$\bullet$ In \textbf{modern logic}, universal propositions lack EI. "All $S$ are $P$" is formalized as a conditional, $\forall x\,(Sx \rightarrow Px)$, which can remain true even if no $S$ exists (i.e., it is \emph{vacuously true}).

We illustrate this contrast with the unicorn example in Figure~\ref{fig:existential import problem}. Under the traditional EI reading, the universal premise ``All unicorns are mammals'' is commonly taken to license the existential conclusion ``Some unicorns are mammals.'' Under modern logic, however, the universal premise entails only $\forall x\,(Ux \rightarrow Mx)$ and does not imply $\exists x\,Ux$; therefore the existential conclusion does not follow unless we add an explicit existence premise.

\subsection{Dataset Construction}
\label{sec:dataset}

We build our data for analysis with a multi-stage agent pipeline that proposes terms and relations, checks factual consistency, and enforces logical constraints before generating syllogistic instances. Using this process, we generate 100 concept triplets with an empty minor-term extension and 100 with a non-empty minor-term extension, combined with Chinese/English versions and 15+9 syllogistic forms, which yields 9600 syllogisms in total. More detailed of the data construction will be discussed in Appendix~\ref{data}.


\begin{table*}[t]
\centering
\resizebox{\textwidth}{!}{%
\begin{tabular}{l|ccc|ccc|ccc|ccc}
\toprule
\textbf{Model} & \multicolumn{3}{c}{\textbf{ZH+}} & \multicolumn{3}{c}{\textbf{ZH-}} & \multicolumn{3}{c}{\textbf{EN+}} & \multicolumn{3}{c}{\textbf{EN-}} \\
\cmidrule(lr){2-4}
\cmidrule(lr){5-7}
\cmidrule(lr){8-10}
\cmidrule(lr){11-13}
  & \textbf{$\text{Acc}_t$} & \textbf{$\text{Acc}_{m}$} & \textbf{$\text{Cons}$} & \textbf{$\text{Acc}_t$} & \textbf{$\text{Acc}_{m}$} & \textbf{$\text{Cons}$} & \textbf{$\text{Acc}_t$} & \textbf{$\text{Acc}_{m}$} & \textbf{$\text{Cons}$} & \textbf{$\text{Acc}_t$} & \textbf{$\text{Acc}_{m}$} & \textbf{$\text{Cons}$} \\
\midrule
\multicolumn{13}{c}{\textbf{Qwen Series -- Dense Models}}\\
\cmidrule(lr){1-13}
Qwen3-0.6B & 100.00 & 62.50 & 100.00 & 99.96 & 62.46 & 95.83 & 100.00 & 62.50 & 100.00 & 100.00 & 62.50 & 100.00 \\
Qwen3-0.6B-Thinking & 94.71 & 61.04 & 4.17 & 92.96 & 61.12 & 16.67 & 86.67 & 60.25 & 0.00 & 88.33 & 61.75 & 4.17 \\
Qwen3-1.7B & 97.00 & 62.42 & 50.00 & 95.58 & 60.92 & 37.50 & 75.21 & 59.71 & 16.67 & 35.17 & 47.58 & 4.17 \\
Qwen3-1.7B-Thinking & 92.92 & 67.67 & 29.17 & 94.29 & 67.71 & 50.00 & 91.62 & 70.54 & 54.17 & 91.96 & 70.29 & 58.33 \\
Qwen3-4B & 92.46 & 67.12 & 45.83 & 94.46 & 67.04 & 54.17 & 85.79 & 61.62 & 4.17 & 93.50 & 61.67 & 12.50 \\
Qwen3-4B-Thinking & 82.54 & 79.96 & 62.50 & 85.33 & 77.08 & 58.33 & 83.62 & 78.88 & 66.67 & 84.92 & 77.58 & 62.50 \\
Qwen3-8B & 94.12 & 67.46 & 33.33 & 96.67 & 65.42 & 62.50 & 85.46 & 69.58 & 4.17 & 86.71 & 64.62 & 0.00 \\
Qwen3-8B-Thinking & 67.83 & 94.50 & 54.17 & 71.62 & 90.88 & 62.50 & 64.83 & 97.67 & 75.00 & 65.29 & 97.21 & 66.67 \\
Qwen3-14B & 97.75 & 64.50 & 66.67 & 99.25 & 63.25 & 87.50 & 87.12 & 70.96 & 25.00 & 91.58 & 68.08 & 20.83 \\
Qwen3-14B-Thinking & 72.96 & 89.54 & 62.50 & 76.50 & 86.00 & 66.67 & 74.92 & 87.50 & 58.33 & 77.92 & 84.50 & 58.33 \\
Qwen3-32B & 91.67 & 70.33 & 58.33 & 95.54 & 66.96 & 75.00 & 91.00 & 70.50 & 45.83 & 93.88 & 68.46 & 54.17 \\
Qwen3-32B-Thinking & 82.21 & 80.29 & 62.50 & 85.75 & 76.75 & 62.50 & 77.96 & 84.50 & 62.50 & 80.38 & 82.08 & 62.50 \\
\cmidrule(lr){1-13}
\multicolumn{13}{c}{\textbf{Qwen Series -- MoE Models}}\\
\cmidrule(lr){1-13}
Qwen3-30B-A3B-Instruct & 66.58 & 95.83 & 70.83 & 71.96 & 90.54 & 66.67 & 64.00 & 98.50 & 75.00 & 66.71 & 95.71 & 66.67 \\
Qwen3-30B-A3B-Thinking & 69.17 & 93.33 & 62.50 & 71.50 & 91.00 & 62.50 & 67.71 & 86.12 & 16.67 & 70.00 & 84.08 & 8.33 \\
Qwen3-NEXT-80B-A3B-Instruct & 65.58 & 96.92 & 66.67 & 70.08 & 92.42 & 66.67 & 62.71 & 99.62 & 70.83 & 64.38 & 98.12 & 62.50 \\
Qwen3-NEXT-80B-A3B-Thinking & 62.71 & 99.79 & 83.33 & 63.08 & 99.42 & 79.17 & 62.88 & 98.96 & 50.00 & 62.96 & 99.38 & 75.00 \\
Qwen3-235B-A22B-Instruct & 66.17 & 96.33 & 66.67 & 67.83 & 94.67 & 66.67 & 62.54 & 99.88 & 87.50 & 62.71 & 99.79 & 83.33 \\
Qwen3-235B-A22B-Thinking & 62.71 & 99.79 & 83.33 & 62.88 & 99.62 & 83.33 & 64.75 & 97.75 & 62.50 & 63.08 & 99.42 & 70.83 \\

\cmidrule(lr){1-13}
\multicolumn{13}{c}{\textbf{Gemma Series}}\\
\cmidrule(lr){1-13}
Gemma-3-1B-IT & 87.96 & 53.29 & 0.00 & 77.62 & 51.71 & 0.00 & 90.29 & 57.54 & 0.00 & 86.71 & 57.54 & 0.00 \\
Gemma-3-4B-IT & 94.46 & 63.38 & 16.67 & 77.88 & 63.54 & 0.00 & 95.00 & 63.08 & 12.50 & 94.79 & 64.38 & 25.00 \\
Gemma-3-12B-IT & 98.54 & 63.38 & 41.67 & 98.96 & 62.88 & 45.83 & 93.67 & 63.42 & 20.83 & 92.38 & 64.96 & 20.83 \\
Gemma-3-27B-IT & 95.33 & 62.00 & 16.67 & 94.17 & 61.58 & 20.83 & 96.54 & 65.71 & 50.00 & 95.96 & 66.54 & 66.67 \\
\cmidrule(lr){1-13}
\multicolumn{13}{c}{\textbf{Llama Series}}\\
\cmidrule(lr){1-13}
Llama3-8B-Instruct & 75.12 & 60.21 & 0.00 & 63.29 & 53.79 & 0.00 & 50.25 & 56.88 & 0.00 & 47.42 & 51.83 & 0.00 \\
Llama3-70B-Instruct & 98.58 & 63.17 & 58.33 & 96.88 & 62.71 & 45.83 & 98.88 & 62.54 & 62.50 & 90.67 & 60.29 & 20.83 \\
Llama3.3-70B-Instruct & 96.08 & 65.92 & 58.33 & 97.88 & 63.96 & 62.50 & 99.08 & 63.00 & 87.50 & 99.12 & 63.38 & 79.17 \\

\cmidrule(lr){1-13}
\multicolumn{13}{c}{\textbf{Closed-source Models}}\\
\cmidrule(lr){1-13}
Claude-3.7-Sonnet  & 85.29 & 76.54 & 45.83 & 90.46 & 71.71 & 50.00 & 70.33 & 92.00 & 54.17 & 73.08 & 89.42 & 62.50 \\
Claude-4.5-Sonnet  & 81.38 & 81.12 & 62.50 & 93.96 & 68.57 & 62.50 & 70.01 & 92.52 & 66.67 & 84.11 & 78.40 & 62.50 \\
Gemini-2.5-Pro & 71.92 & 89.33 & 29.17 & 76.17 & 83.50 & 25.00 & 65.17 & 97.33 & 70.83 & 72.92 & 89.50 & 58.33 \\
Gemini-3-Pro-Preview & 73.11 & 89.20 & 54.17 & 99.00 & 63.48 & 66.67 & 63.48 & 99.00 & 79.17 & 98.41 & 64.02 & 70.83 \\
GPT-4o-2024-11-20  & 93.17 & 68.42 & 41.67 & 96.17 & 65.71 & 50.00 & 93.33 & 68.75 & 50.00 & 94.04 & 67.83 & 50.00 \\
GPT-4.1-2025-04-14  & 80.38 & 80.04 & 33.33 & 85.08 & 76.67 & 45.83 & 80.04 & 82.38 & 58.33 & 81.54 & 80.96 & 62.50 \\
GPT-o3 & 62.38 & 99.54 & 87.50 & 62.58 & 99.92 & 91.67 & 62.50 & 100.00 & 100.00 & 62.58 & 99.92 & 95.83 \\
GPT-5-2025-08-07 & 62.50 & 100.00 & 100.00 & 62.50 & 100.00 & 100.00 & 62.50 & 100.00 & 100.00 & 62.50 & 100.00 & 100.00 \\
\bottomrule
\end{tabular}
}
\caption{Results for various models by language and the subject term's existence condition (non-empty vs.\ empty extension). Detailed metrics (e.g., precision and recall) are reported in the Appendix ~\ref{sec:appendix_main_results}.}
\label{tab:all_models_by_lang_concept_brief}
\end{table*}

\section{Experiment Design}
\subsection{The 15+9 Distinction of Valid Syllogistic Forms}
This disagreement over EI directly creates a split in the set of valid syllogistic forms. A form is defined by its mood (the A/E/I/O pattern) and figure (term arrangement).
\begin{itemize}
    \item \textbf{Traditional Logic} recognizes 24 valid forms.
    \item \textbf{Modern Logic} accepts only 15 of these as unconditionally valid. The remaining 9 forms are rejected precisely because they commit the existential fallacy.
\end{itemize}
As shown in Appendix~\ref{24valid}, we use 15+9 split to distinguish traditional from modern logic validity, and report accuracy under each logic paradigm accordingly.

We further compare a baseline prompt with a \textsc{Prior-check} prompt that explicitly asks the model to first state whether the concepts are empty in the given setting by add \textit{"Do you think \{major term\}, \{middle term\}, \{minor term\} are empty sets? Keep that in mind and answer:"} at the beginning of prompt, testing whether making the existence status explicit shifts the model’s behavior between traditional and modern logic.




We evaluate model behavior under both traditional logic and modern logic, and also examine how stable its reasoning is across instances of the same syllogistic form. 

We first report \textbf{traditional-logic accuracy ($\text{Acc}_t$)}, defined as the proportion of instances in which the model accepts the conclusion, treating all 24 moods as valid under existential import. We then report \textbf{modern-logic accuracy ($\text{Acc}_{m}$)}, defined with respect to modern semantics: the model should accept instances from the 15 moods that are valid in modern logic, and reject instances from the 9 moods that become invalid when the minor term $S$ has an empty extension. Higher 
$\text{Acc}_t$ indicates behavior closer to traditional logic, while higher $\text{Acc}_{m}$ indicates behavior more consistent with the modern logic. 

Moreover, \textbf{consistency score ($\text{Cons}$)} of each mood in each language and concept-emptiness set is report as $\frac{n}{24}$. Model can earn the score only if all answers of the same mood is consistent. In addition, we report \textbf{precision and recall} separately on the two mood subsets (the 15 unconditionally-valid moods and the 9 existential-import-dependent moods) to better characterize how the model distinguishes between these two logic regimes， detailed in Appendix~\ref{precision_recall}.

\section{Results and Analysis}
\label{section4}

\subsection{Main Results}

\subsubsection{Scaling Effects of Logical Evolution}

\paragraph{Advanced models exhibit modern-logic behavior.}
Widely recognized as advanced closed-source LLMs(e.g., Gemini-2.5-Pro \cite{comanici2025gemini25}, GPT-o3 \cite{openai_gpto3}, GPT-5 \cite{openai_gpt5}) increasingly prefer modern logic while maintaining relatively low scores under the traditional logic (see Table ~\ref{tab:all_models_by_lang_concept_brief}). This change is not only about higher accuracy, but also suggests that models are moving toward a more rule-based and principled way to analyze validity.

Motivated by this, we ask a basic question: \textit{how does the preference for the modern logic emerge as models are developed and scaled up?} To study this in a controlled way, we turn to open-source model families where we can compare many related checkpoints. 
Concretely, we evaluate the Qwen \cite{yang2025qwen3technicalreport}, Llama \cite{grattafiori2024llama3herdmodels}, and Gemma \cite{gemmateam2025gemma3technicalreport} series. Overall, we find that as model size increases, $Acc_m$ rises across all models, indicating that models' logical reasoning became more rigorous as the parameter scaling up.


\paragraph{Clear family-specific scaling patterns.}
We further conduct a detailed analysis of three major model families. Since Qwen series provide comprehensive coverage across the wide scale range from 0.6B to 235B, multiple variants, and different architectures (including dense and mixture-of-experts models), we primarily analyze Qwen models and report them as our main results.



Among these three families, we find clear family-specific scaling patterns in logical behavior.
Qwen shows a scaling trend that includes \textbf{a clear logic paradigm shift}.
For small to mid-sized non-thinking and instruction-tuned Qwen models, $\text{Acc}_t$ remains very high, indicating a strong preference for the traditional logic.
However, when moving to larger Qwen models---especially thinking variants and some large instruction-tuned variants---the pattern can flip, with $\text{Acc}_{m}$ becoming much higher than $\text{Acc}_t$.
This trend holds in both Chinese and English, suggesting it is not tied to a single language setting.
In contrast, for Llama and Gemma, models at different sizes mostly follow the traditional logic.
Scaling mainly makes them stronger within the traditional logic. 

We hypothesize this is because, at small sizes, the model gradually grasps traditional logic to improve inference performance. However, at larger sizes, to solve more complex problems, the model must switch to full modern logic.
We also observe the consistency scores fluctuate when scaling up Qwen models.
This instability is more likely near the \textbf{Turning Point} where the model's logic switches.
This suggests the transition from traditional logic to modern logic is not always smooth. During the change, the model may mix the behaviors of following surface patterns from data and more rigorous reasoning of modern logic, which can temporarily lead to disagreements across closely related test cases.

\begin{AIbox}{Takeaway 1}
As models scale up, their logic judgments clearly shift from the traditional logic to modern logic, matching the same direction we see in advanced closed-source models.
\end{AIbox}

\subsubsection{Thinking as an Efficient Driver of the Logic Evolution}

\paragraph{Thinking accelerates the logic shift at fixed scale.}
Since Thinking directly strengthens a model’s multi-step reasoning process, it enables more consistent rule-based inference with less reliance on scale alone. We compare same-sized \texttt{Instruct/Non-thinking} models with their \texttt{Thinking} counterparts. The results show that the thinking mechanism can strongly speed up the shift from the traditional logic to the modern logic.
This is most obvious in the Qwen3-8B pair: while Qwen3-8B still mostly follows the traditional logic, Qwen3-8B-Thinking moves clearly toward the modern logic stance.
For larger models where the \texttt{Instruct} version is already strongly modern-logic-aligned, the \texttt{Thinking} version often further improves $\text{Acc}_{m}$ and increases consistency across closely related test cases.

A natural explanation is that reinforcement learning (RL) makes the model rely more on step-by-step, rule-like deduction, and also helps it give more stable answers when two cases are very similar.
In this sense, \texttt{Thinking} does not just add better instruction following. It changes the decision criterion and makes the logic paradigm shift more likely.

\paragraph{Thinking is an efficient alternative to parameter scaling.}
Under the modern logic, Qwen3-8B-Thinking can reach a performance level close to Qwen3-30B-A3B-Instruct, even though it uses far fewer parameters.
So, increasing model size is not the only way to get strong modern logic behavior. RL training with explicit reasoning traces can partly replace the need for more parameters by changing how the model uses its capacity.
In practice, scaling tends to improve broad robustness but is expensive, while RL-based thinking can be a more focused and compute-efficient way to push the model into the modern logic.
The best results still come from combining large model further enhanced with RL.

\paragraph{CoT Prompting and Distillation are insufficient.}
To further investigate the effectiveness of \texttt{Thinking} mechanism, we conduct two additional experiments. First, starting from the \texttt{Instruct} models, we add an explicit CoT-trigger prompt (e.g., \textsf{"Let's think step by step."}). The results are reported in Appendix~\ref{thinking_external}.
We find that \texttt{Instruct+CoT} setting can induce a partial shift toward modern logic, but the shift is limited.
In contrast, the \texttt{Thinking} models produce a more complete transition in the underlying logic criterion, further supporting our main finding that RL-trained thinking acts as a promoter of logic shift.

In addition, we also examine several distilled models derived from large RL-trained model (e.g., DeepSeek-R1-Distill-Llama-8B \cite{deepseekai2025deepseekr1}). The results in Appendix~\ref{thinking_external} suggest that RL training does not automatically lead to rigorous modern logic in all models. Instead, achieving a stable shift to modern logic appears to require careful, task-aware design. At least in our setting, distillation from DeepSeek-R1 alone is far from sufficient to produce the same level of strict modern logic behavior.

\begin{AIbox}{Takeaway 2}
The thinking process derived from RL can push a smaller model into modern logic.
\end{AIbox}

\subsubsection{Base Models as the Starting Point and a Constraint}

\paragraph{Base models shape what post-training can achieve.}
Scaling and RL can change a model's logical stance, but these changes do not start from nowhere.
Here we test a more basic point: \textit{how much the final behavior is already shaped by the underlying Base model?}
To answer this, we evaluate several base models (Appendix~\ref{base_model}).
Overall, the base model sets \textbf{the starting point} for post-training, and it strongly affects both (i) what the later \texttt{Instruct}/\texttt{Thinking} models can learn and (ii) how stable that learned criterion will be.

\paragraph{Modern-logic signals at the Base stage enable easier shifts.}
From Qwen3-8B-Base, where we later observe a clear shift toward modern logic, we already see an important signal at the base stage.
It achieves relatively high $rec_V$, and in most settings its $rec_I$ is also clearly higher than other base models.
This suggests that Qwen3-8B-Base is not fully locked by traditional logic.
Instead, it already shows some ability to separate the modern-valid moods from all moods, leaving room for post-training to strengthen modern logic.
This explains why RL in the \texttt{Thinking} variant can push Qwen3-8B toward modern more easily.

In contrast, Gemma and LLaMA Base models often have low $rec_V$, meaning they frequently fail to recognize modern-valid moods and tend to answer "invalid" by default.
This also explains their seemingly high $rec_I$ on the existential-import-dependent subset: the high $rec_I$ is largely caused by a general rejection tendency, rather than real sensitivity to existential import.

\paragraph{The effect of Base model is strong but not absolute .}
Small models (e.g., Qwen3-8B) benefit the most when the Base already shows modern logic signals.
Larger models can still learn the modern logic through post-training (e.g., \texttt{Qwen3-30B-A3B}), but the learned shift is not always stable: under \texttt{Thinking}, judgments can fluctuate, and in some cases the model can drift back toward a traditional pattern.
This suggests that post-training can move the decision criterion, but the base model still influences how reliable that move will be.

\begin{AIbox}{Takeaway 3}
The base model is the starting point. If it already leans toward modern logic, post-training shifts are easier and more stable.
\end{AIbox}

\begin{figure*}[t]
  \centering
  \captionsetup[subfigure]{font=scriptsize} 
  \begin{subfigure}[t]{0.24\textwidth}
    \centering 
    \includegraphics[width=\linewidth]{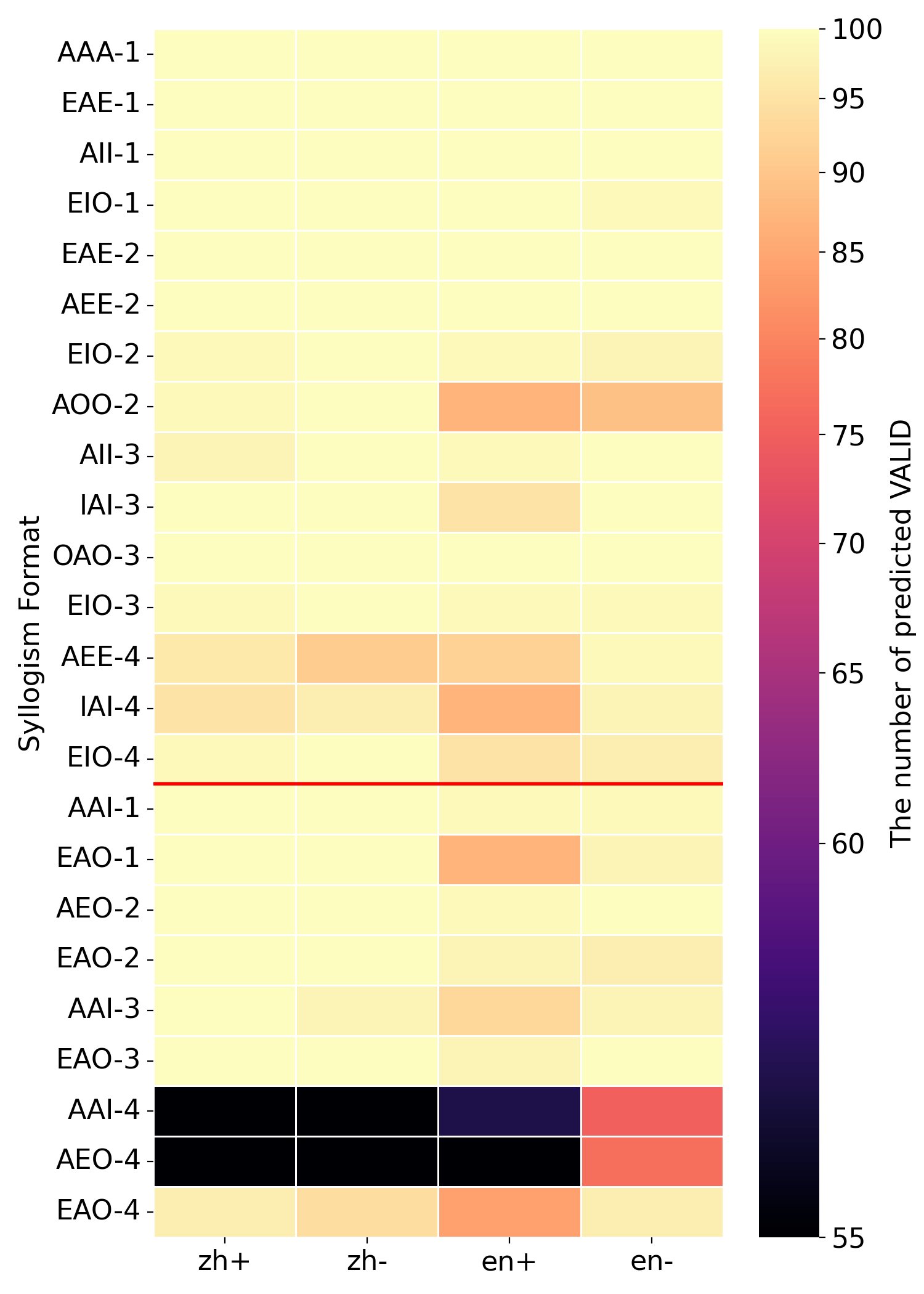}
    \caption{Qwen3-4B}
  \end{subfigure}\hfill
  \begin{subfigure}[t]{0.24\textwidth}
    \centering
    \includegraphics[width=\linewidth]{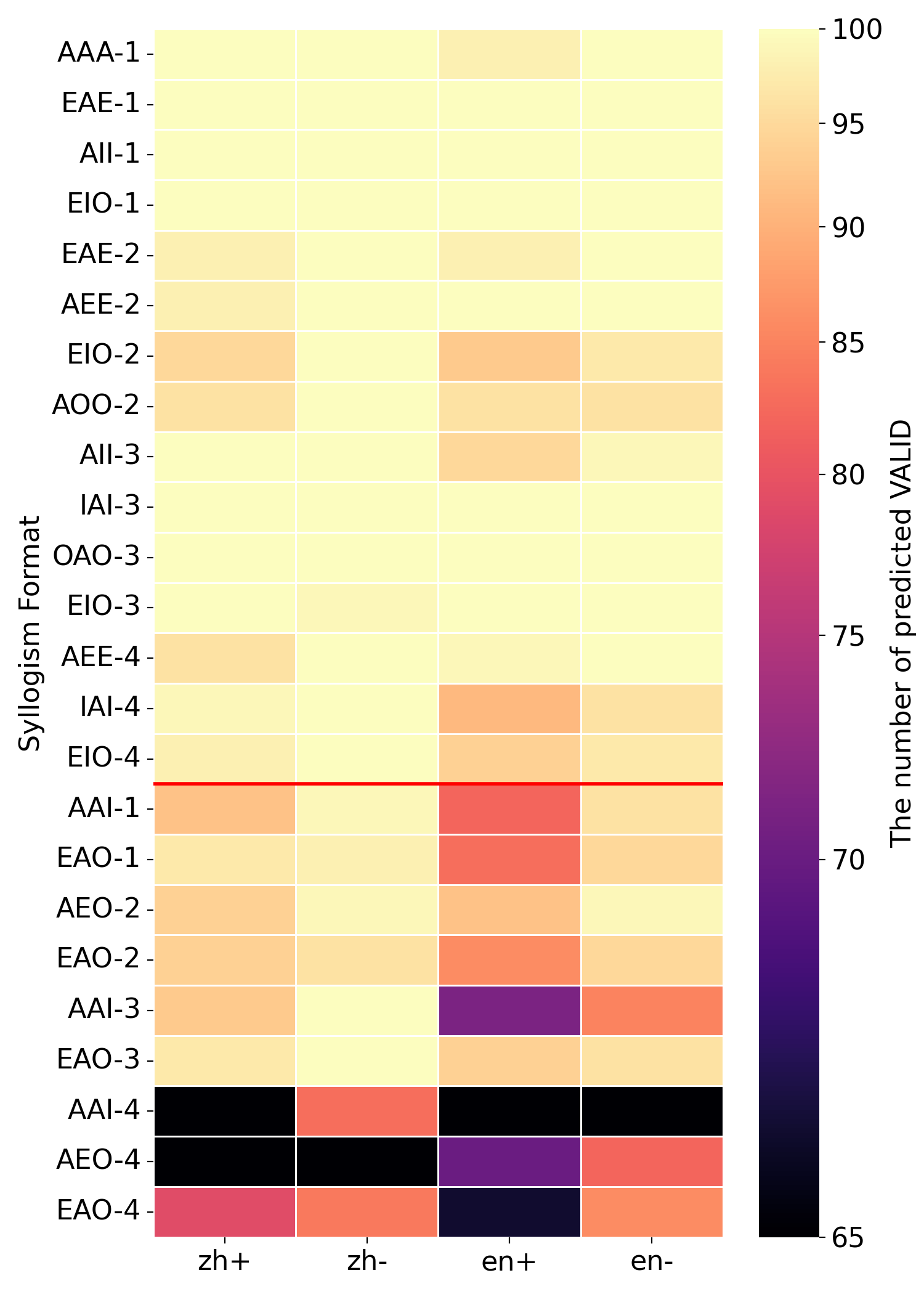}
    \caption{Qwen3-8B}
  \end{subfigure}\hfill
  \begin{subfigure}[t]{0.24\textwidth}
    \centering
    \includegraphics[width=\linewidth]{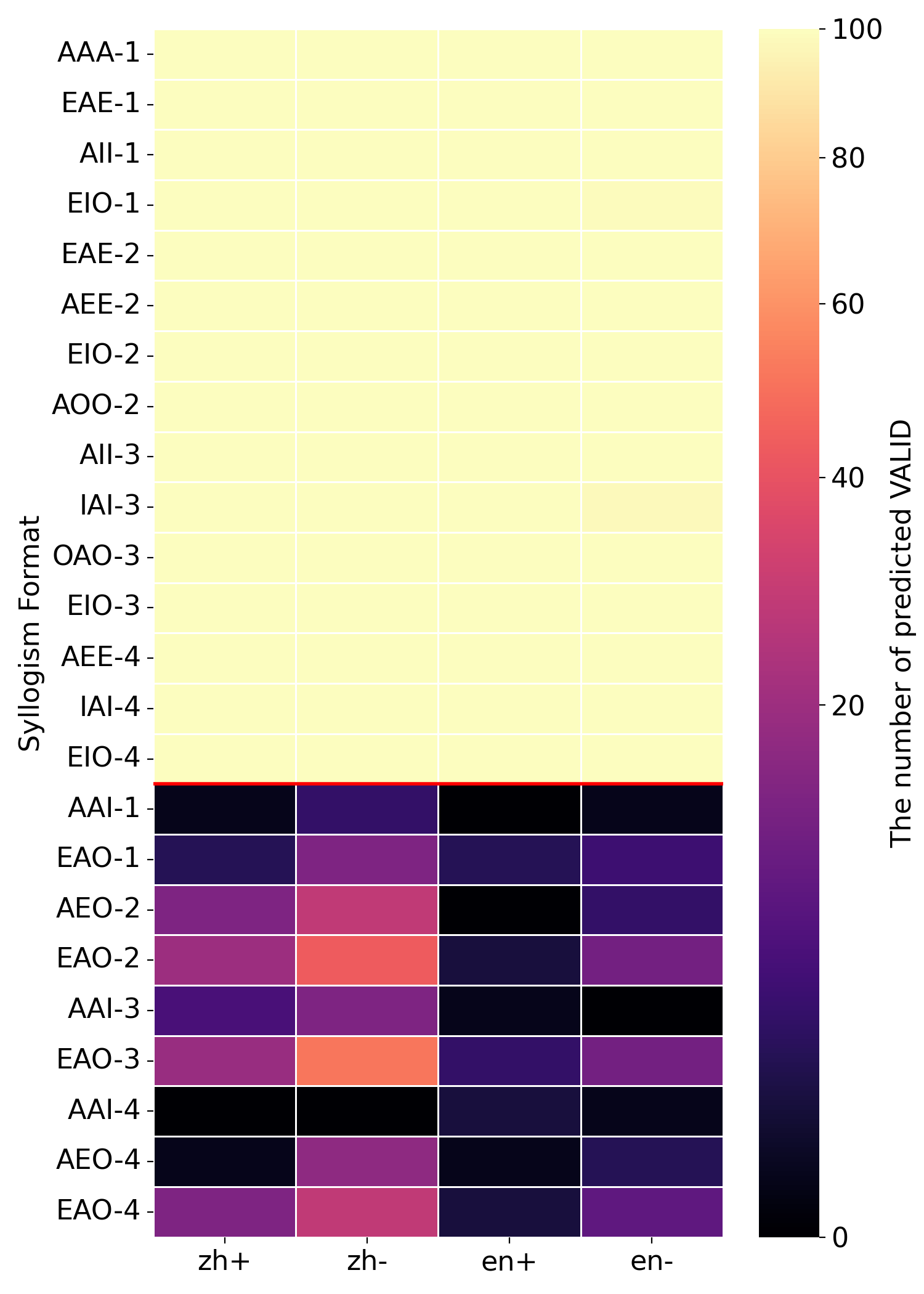}
    \caption{Qwen3-NEXT-80B-A3B-Instruct}
  \end{subfigure}\hfill
  \begin{subfigure}[t]{0.24\textwidth}
    \centering
    \includegraphics[width=\linewidth]{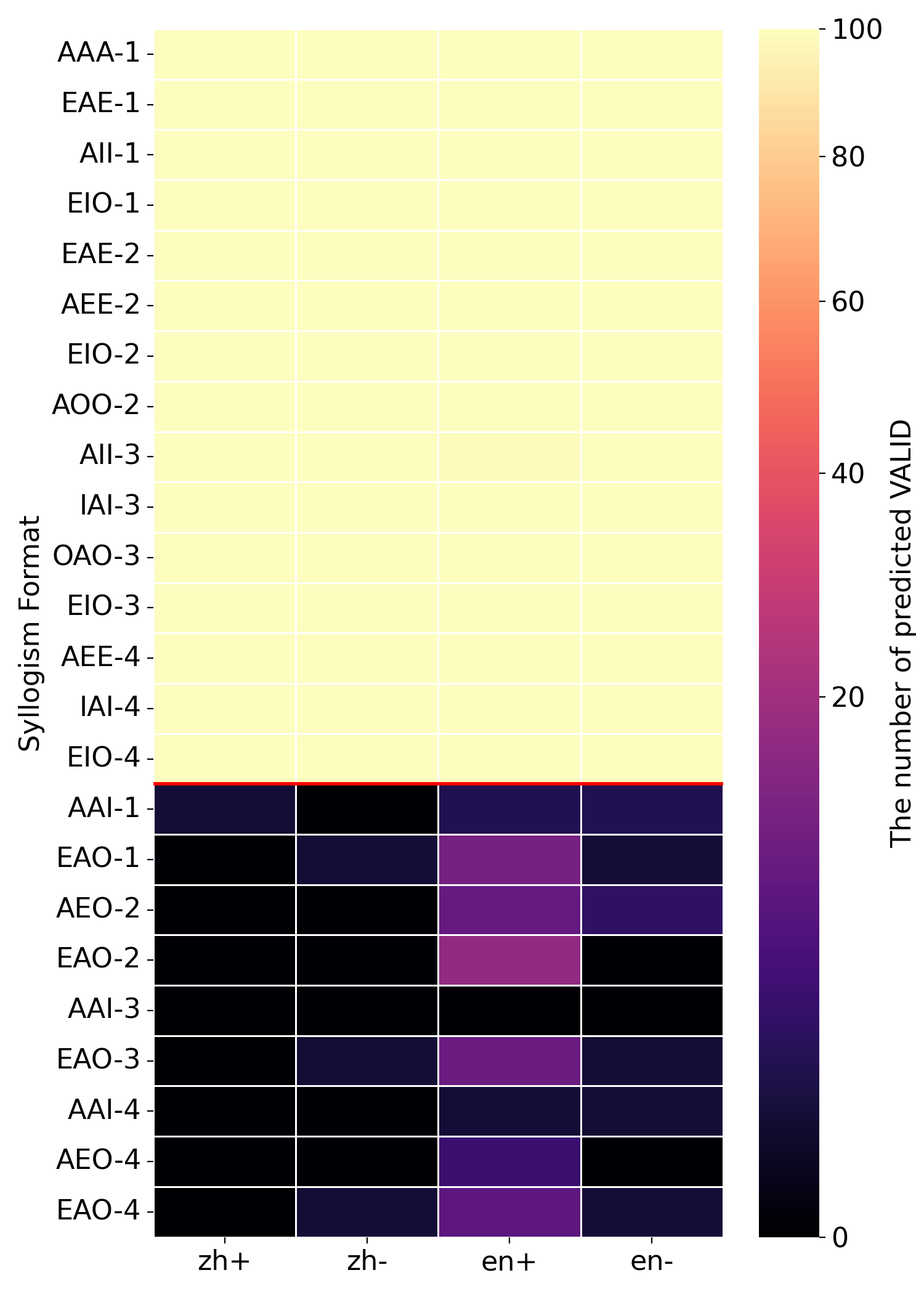}
    \caption{Qwen3-235B-A22B-Thinking}
  \end{subfigure}
  \caption{The heatmaps of two types of model logic. (a) and (b) are traditional logic while (c) and (d) are modern logic.}
\label{fig:heatmaps}
\end{figure*}

\subsection{Further Analysis}

\subsubsection{Prior-check Prompt}

To ensure that our measured modern logic performance is not an artifact of prompting, we introduce \emph{Prior-check} prompt that explicitly asks the model to check the relevant existence condition before making a validity judgment. The goal is simple: make the model perform a semantic check that is required under modern logic evaluation, without changing the logical content of the task.

\paragraph{Main effect: higher $\mathrm{Acc}_m$ without stance flipping.}
As a control group, we report results with baseline prompt in Appendix~\ref{sec:appendix_control_group}. We observe that Prior-check prompt consistently increases $\text{Acc}_{m}$ for most models, while keeping their overall logical stance stable and easy to interpret.
This suggests that the prompt improves compliance with modern logic rather than introducing systematic bias.

\paragraph{Turning-point instability.}
A notable exception appears in the Qwen3-30B-A3B pair. Although the \texttt{Instruct} version looks modern-logic, the \texttt{Thinking} version shifts back toward traditional logic. This suggests that Qwen3-30B-A3B model is close to the turning point between paradigms. Long thinking contents may sometimes bring back traditional defaults. The fluctuations reveal that the model's stance can be fragile during the logic transition stage.

\subsubsection{The emptiness of minor term}

\paragraph{Empty minor terms are consistently harder.}
Under both Prior-check prompt and the baseline setting, models show lower $rec_I$ when the minor term is empty than when the non-empty counterpart.

One likely reason is that empty minor terms make counterexamples harder to construct. To judge an argument as invalid under modern logic, the model often needs to consider a situation where the premises are true but the conclusion is false. When the minor term is empty, this kind of reasoning is less intuitive because there are no concrete instances to reason about. As a result, the model tends to fall back on traditional logic. This increases false positives and reduces $rec_I$. This result highlights the permeability of world knowledge. Plausibility priors can leak into formal reasoning and interfere with rule-governed validity judgments.

\paragraph{Mood-specific error concentration suggests data imprinting.}
To further probe knowledge effects in syllogistic reasoning, we visualize the number of "valid" answers across languages, minor-term existence settings, and all 24 syllogistic moods, as shown in Figure~\ref{fig:heatmaps}. The figure compares four models under Prior-check prompt. Regardless of whether a model generally aligns with traditional or modern logic,  errors concentrate on a few specific moods rather than being evenly distributed. For example, Qwen3-4B is overall closer to traditional logic, it displays a strong tendency toward the modern logic in the AAI-4 and AEO-4 syllogism forms. 
One explanation is that certain moods are more frequent in training data, leading to better learning of those forms. This supports the view that LLMs’ logical behavior is shaped by training data, rather than reflecting an abstract reasoning ability that generalizes uniformly.


\subsubsection{Cross-lingual Gaps}

\textbf{Clear language-dependent effect.} When comparing three open-source series, the Qwen and LLaMA series generally perform better in Chinese than in English, while Gemma shows the opposite pattern, with higher performance in English. This difference is most visible in accuracy measured under each model’s dominant logical stance.

This cross-lingual gap suggests that current LLMs’ logical ability is not fully language-agnostic. Instead, it is still strongly shaped by language-specific patterns in training data. In short, what looks like “logical reasoning” in these models is still partly tied to the language they operate in, rather than being a truly language-independent reasoning skill.

\subsubsection{Architecture and Reasoning Ability}

We next study how model architecture relates to logical reasoning behavior. We consider two settings: (i) open-source auto-regressive (AR) LLMs, comparing Dense models with mixture-of-experts (MoE) models shown in Table~\ref{tab:all_models_by_lang_concept_brief}; and (ii) emerging diffusion LLMs (dLLMs) shown in Table~\ref{tab:dllm_by_lang_concept}.

\paragraph{MoE in AR models correlates with more modern-leaning behavior.}
Within AR models, MoE variants in the Qwen family exhibit a stronger tendency toward modern logic than same-generation dense models. A plausible explanation is the combined effect of MoE efficiency and model scaling. MoE architectures make it easier to train models with higher effective capacity under similar compute, and the shift toward modern logic becomes more likely as model size increases.

\paragraph{DLLMs mostly follow traditional logic.}
For dLLMs, most models still predominantly follow the traditional logic. Only one exception is LLaDA2.0-flash, which is a 100B model with MoE architecture. This exception again reflects the joint impact of MoE architecture and model scaling.

\section{Related Works}

In recent years, many benchmarks have been proposed for syllogism reasoning. ENN \cite{dong2020enn} constructed syllogisms extracted from WordNet \cite{wordnet}. The syllogsims are in the form of triplets with no natural language descriptions. Syllo-Figure \cite{Syllogism_Figure} and NeuBAROCO \cite{ando2023NeuBAROCO} are two natural language syllogism datasets, with data derived from existed datasets. Syllo-Figure derives omitted syllogisms from SNLI \cite{bowman2015largeannotatedcorpuslearning} and rewrites the missing premise by annotators. The target is to identify the specific figure. NeuBAROCO transforms questions from BAROCO \cite{SHIKISHIMA2009256} into a format used for natural language inferences(NLI). Beyond categorical syllogism, SylloBase \cite{wu-etal-2023-hence} covers more types and patterns of syllogism, covering  a complete taxonomy of syllogism reasoning patterns. There are also several researches focusing on the human-like bias of syllogism, such as belief bias \cite{nguyen2025bisreasoning, ando2023NeuBAROCO} and atmosphere effects \cite{ando2023NeuBAROCO}.
However, these works all assume existential import by default, meaning they approach the task under a traditional logic setting. To examine different models’ tendencies under different logical paradigms and gain deeper insights, we use existential import as a probe and conduct a series of investigations.

\section{Conclusion and Discussion}
\label{section6}

This work studies whether LLMs’ syllogistic validity judgments shift toward a more rigorous modern logic criterion as models develop. Among all models, $\mathrm{Acc}_m$ generally increases with scale, but only the Qwen series exhibits a clear logic shift , consistent with the behavior of advanced closed-source models. Matched-size comparisons further show that RL-trained \texttt{Thinking} variants efficiently accelerate this shift and improve consistency; in contrast, CoT prompting induces only a limited move toward modern logic, and distillation alone does not reliably yield strict modern logic behavior.

However, the transition is not always smooth. The consistency can fluctuate near the turning point, and some near-boundary models (e.g., Qwen3-30B-A3B) may partially revert under reasoning traces. We also identify systematic failure modes that persist across settings, including difficulty with empty minor terms, mood-specific bias, and cross-lingual gaps. Overall, our results suggest that modern logic reasoning in LLMs is shaped jointly by the base model and post-training (especially RL-based thinking), rather than emerging from parameter scaling alone.

\section*{Limitations}

Our conclusions are primarily drawn from syllogistic reasoning and the contrast induced by existential import. While this probe cleanly separates traditional and modern validity criteria, it remains unclear whether the same evolutionary patterns hold for broader first-order logic criteria.

We evaluate models mainly through their final valid/invalid decisions. This endpoint-only metric can obscure the source of errors. Our study does not directly supervise or diagnose intermediate semantic representations or proof-like structures, limiting our ability to pinpoint the mechanisms behind observed shifts and inconsistencies.

Our distillation analysis covers only a small set of distilled models and a specific teacher family (e.g., DeepSeek-R1). Moreover, the distillation objectives and data are not fully known or comparable across models. As a result, our finding that distillation alone does not reliably induce strict modern-logic behavior should be interpreted as an empirical observation in our setting, rather than a general negative result about distillation.


\bibliography{custom}

@article{rasooli-tetrault-2015,
    author    = {Mohammad Sadegh Rasooli and Joel R. Tetreault},
    title     = {Yara Parser: {A} Fast and Accurate Dependency Parser},
    journal   = {Computing Research Repository},
    volume    = {arXiv:1503.06733},
    year      = {2015},
    url       = {http://arxiv.org/abs/1503.06733},
    note    = {version 2}
}

@incollection{Aristotle,
  author    = {Aristotle},
  editor    = {Barnes, Jonathan},
  title     = {Prior Analytics},
  booktitle = {The Complete Works of Aristotle: The Revised Oxford Translation, Vol. 1},
  publisher = {Princeton University Press},
  address   = {Princeton, NJ},
  year      = {1984},
  pages     = {39--113},
}

@book{Copi2014Logic,
  author    = {Copi, Irving M. and Cohen, Carl and McMahon, Kenneth},
  title     = {Introduction to Logic},
  edition   = {14},
  publisher = {Pearson Education},
  address   = {Harlow, England},
  year      = {2014}
}

@InCollection{sep-square,
	author       =	{Parsons, Terence and Ciola, Graziana},
	title        =	{{The Traditional Square of Opposition}},
	booktitle    =	{The {Stanford} Encyclopedia of Philosophy},
	editor       =	{Edward N. Zalta and Uri Nodelman},
	howpublished =	{\url{https://plato.stanford.edu/archives/sum2025/entries/square/}},
	year         =	{2025},
	edition      =	{{S}ummer 2025},
	publisher    =	{Metaphysics Research Lab, Stanford University}
}

@book{Suppes1957-SUPITL,
	address = {Mineola, N.Y.},
	author = {Patrick Suppes},
	editor = {},
	publisher = {Dover Publications},
	title = {Introduction to Logic},
	year = {1957}
}

@book{Boole1854,
  author    = {Boole, George},
  title     = {An Investigation of the Laws of Thought: On Which Are Founded the Mathematical Theories of Logic and Probabilities},
  publisher = {Walton and Maberly},
  address   = {London},
  year      = {1854}
}

@book{enderton-math-logic,
  author    = {Enderton, Herbert B.},
  title     = {A Mathematical Introduction to Logic},
  year      = {1972},
  publisher = {Academic Press},
  address   = {New York}
}

@online{openai_gpt5,
  author  = {{OpenAI}},
  title   = {Introducing GPT-5},
  year    = {2025},
  url     = {https://openai.com/index/introducing-gpt-5/},
  urldate = {2026-01-05},
}

@online{openai_gpto3,
  author  = {{OpenAI}},
  title   = {Introducing OpenAI o3 and o4-mini},
  year    = {2025},
  url     = {https://openai.com/index/introducing-o3-and-o4-mini/},
  urldate = {2026-01-05},
}

@online{gemini_3_pro,
  author  = {{deepmind}},
  title   = {Gemini 3 Pro
Best for complex tasks and bringing creative concepts to life},
  year    = {2025},
  url     = {https://deepmind.google/models/gemini/pro/},
  urldate = {2026-01-05},
}

@inproceedings{Syllogism_Figure,
author = {Peng, Shiya and Liu, Lu and Liu, Chang and Yu, Dong},
title = {Exploring Reasoning Schemes: A Dataset for Syllogism Figure Identification},
year = {2020},
isbn = {978-3-030-81196-9},
publisher = {Springer-Verlag},
address = {Berlin, Heidelberg},
url = {https://doi.org/10.1007/978-3-030-81197-6_37},
doi = {10.1007/978-3-030-81197-6_37},
booktitle = {Chinese Lexical Semantics: 21st Workshop, CLSW 2020,  Hong Kong, China, May 28–30, 2020,  Revised Selected Papers},
pages = {445–451},
numpages = {7}
}

@article{wordnet,
author = {Miller, George A.},
title = {WordNet: a lexical database for English},
year = {1995},
issue_date = {Nov. 1995},
publisher = {Association for Computing Machinery},
address = {New York, NY, USA},
volume = {38},
number = {11},
issn = {0001-0782},
url = {https://doi.org/10.1145/219717.219748},
doi = {10.1145/219717.219748},
journal = {Commun. ACM},
month = nov,
pages = {39–41},
numpages = {3}
}

@misc{dong2020enn,
      title={Learning Syllogism with Euler Neural-Networks}, 
      author={Tiansi Dong and Chengjiang Li and Christian Bauckhage and Juanzi Li and Stefan Wrobel and Armin B. Cremers},
      year={2020},
      eprint={2007.07320},
      archivePrefix={arXiv},
      primaryClass={cs.LG},
      url={https://arxiv.org/abs/2007.07320}, 
}

@misc{bie2025llada20,
      title={LLaDA2.0: Scaling Up Diffusion Language Models to 100B}, 
      author={Tiwei Bie and Maosong Cao and Kun Chen and Lun Du and Mingliang Gong and Zhuochen Gong and Yanmei Gu and Jiaqi Hu and Zenan Huang and Zhenzhong Lan and Chengxi Li and Chongxuan Li and Jianguo Li and Zehuan Li and Huabin Liu and Lin Liu and Guoshan Lu and Xiaocheng Lu and Yuxin Ma and Jianfeng Tan and Lanning Wei and Ji-Rong Wen and Yipeng Xing and Xiaolu Zhang and Junbo Zhao and Da Zheng and Jun Zhou and Junlin Zhou and Zhanchao Zhou and Liwang Zhu and Yihong Zhuang},
      year={2025},
      eprint={2512.15745},
      archivePrefix={arXiv},
      primaryClass={cs.LG},
      url={https://arxiv.org/abs/2512.15745}, 
}

@misc{cheng2025sdar,
      title={SDAR: A Synergistic Diffusion-AutoRegression Paradigm for Scalable Sequence Generation}, 
      author={Shuang Cheng and Yihan Bian and Dawei Liu and Linfeng Zhang and Qian Yao and Zhongbo Tian and Wenhai Wang and Qipeng Guo and Kai Chen and Biqing Qi and Bowen Zhou},
      year={2025},
      eprint={2510.06303},
      archivePrefix={arXiv},
      primaryClass={cs.LG},
      url={https://arxiv.org/abs/2510.06303}, 
}

@misc{zhu2025dirl,
      title={DiRL: An Efficient Post-Training Framework for Diffusion Language Models}, 
      author={Ying Zhu and Jiaxin Wan and Xiaoran Liu and Siyanag He and Qiqi Wang and Xu Guo and Tianyi Liang and Zengfeng Huang and Ziwei He and Xipeng Qiu},
      year={2025},
      eprint={2512.22234},
      archivePrefix={arXiv},
      primaryClass={cs.LG},
      url={https://arxiv.org/abs/2512.22234}, 
}

@misc{wang2025revolutionizing,
      title={Revolutionizing Reinforcement Learning Framework for Diffusion Large Language Models}, 
      author={Yinjie Wang and Ling Yang and Bowen Li and Ye Tian and Ke Shen and Mengdi Wang},
      year={2025},
      eprint={2509.06949},
      archivePrefix={arXiv},
      primaryClass={cs.CL},
      url={https://arxiv.org/abs/2509.06949}, 
}

@misc{zhu2025llada15variancereducedpreference,
      title={LLaDA 1.5: Variance-Reduced Preference Optimization for Large Language Diffusion Models}, 
      author={Fengqi Zhu and Rongzhen Wang and Shen Nie and Xiaolu Zhang and Chunwei Wu and Jun Hu and Jun Zhou and Jianfei Chen and Yankai Lin and Ji-Rong Wen and Chongxuan Li},
      year={2025},
      eprint={2505.19223},
      archivePrefix={arXiv},
      primaryClass={cs.LG},
      url={https://arxiv.org/abs/2505.19223}, 
}

@misc{nie2025largelanguagediffusionmodels,
      title={Large Language Diffusion Models}, 
      author={Shen Nie and Fengqi Zhu and Zebin You and Xiaolu Zhang and Jingyang Ou and Jun Hu and Jun Zhou and Yankai Lin and Ji-Rong Wen and Chongxuan Li},
      year={2025},
      eprint={2502.09992},
      archivePrefix={arXiv},
      primaryClass={cs.CL},
      url={https://arxiv.org/abs/2502.09992}, 
}

@misc{deepseekai2025deepseekr1,
      title={DeepSeek-R1: Incentivizing Reasoning Capability in LLMs via Reinforcement Learning}, 
      author={DeepSeek-AI and Daya Guo and Dejian Yang and Haowei Zhang and Junxiao Song and Ruoyu Zhang and Runxin Xu and Qihao Zhu and Shirong Ma and Peiyi Wang and Xiao Bi and Xiaokang Zhang and Xingkai Yu and Yu Wu and Z. F. Wu and Zhibin Gou and Zhihong Shao and Zhuoshu Li and Ziyi Gao and Aixin Liu and Bing Xue and Bingxuan Wang and Bochao Wu and Bei Feng and Chengda Lu and Chenggang Zhao and Chengqi Deng and Chenyu Zhang and Chong Ruan and Damai Dai and Deli Chen and Dongjie Ji and Erhang Li and Fangyun Lin and Fucong Dai and Fuli Luo and Guangbo Hao and Guanting Chen and Guowei Li and H. Zhang and Han Bao and Hanwei Xu and Haocheng Wang and Honghui Ding and Huajian Xin and Huazuo Gao and Hui Qu and Hui Li and Jianzhong Guo and Jiashi Li and Jiawei Wang and Jingchang Chen and Jingyang Yuan and Junjie Qiu and Junlong Li and J. L. Cai and Jiaqi Ni and Jian Liang and Jin Chen and Kai Dong and Kai Hu and Kaige Gao and Kang Guan and Kexin Huang and Kuai Yu and Lean Wang and Lecong Zhang and Liang Zhao and Litong Wang and Liyue Zhang and Lei Xu and Leyi Xia and Mingchuan Zhang and Minghua Zhang and Minghui Tang and Meng Li and Miaojun Wang and Mingming Li and Ning Tian and Panpan Huang and Peng Zhang and Qiancheng Wang and Qinyu Chen and Qiushi Du and Ruiqi Ge and Ruisong Zhang and Ruizhe Pan and Runji Wang and R. J. Chen and R. L. Jin and Ruyi Chen and Shanghao Lu and Shangyan Zhou and Shanhuang Chen and Shengfeng Ye and Shiyu Wang and Shuiping Yu and Shunfeng Zhou and Shuting Pan and S. S. Li and Shuang Zhou and Shaoqing Wu and Shengfeng Ye and Tao Yun and Tian Pei and Tianyu Sun and T. Wang and Wangding Zeng and Wanjia Zhao and Wen Liu and Wenfeng Liang and Wenjun Gao and Wenqin Yu and Wentao Zhang and W. L. Xiao and Wei An and Xiaodong Liu and Xiaohan Wang and Xiaokang Chen and Xiaotao Nie and Xin Cheng and Xin Liu and Xin Xie and Xingchao Liu and Xinyu Yang and Xinyuan Li and Xuecheng Su and Xuheng Lin and X. Q. Li and Xiangyue Jin and Xiaojin Shen and Xiaosha Chen and Xiaowen Sun and Xiaoxiang Wang and Xinnan Song and Xinyi Zhou and Xianzu Wang and Xinxia Shan and Y. K. Li and Y. Q. Wang and Y. X. Wei and Yang Zhang and Yanhong Xu and Yao Li and Yao Zhao and Yaofeng Sun and Yaohui Wang and Yi Yu and Yichao Zhang and Yifan Shi and Yiliang Xiong and Ying He and Yishi Piao and Yisong Wang and Yixuan Tan and Yiyang Ma and Yiyuan Liu and Yongqiang Guo and Yuan Ou and Yuduan Wang and Yue Gong and Yuheng Zou and Yujia He and Yunfan Xiong and Yuxiang Luo and Yuxiang You and Yuxuan Liu and Yuyang Zhou and Y. X. Zhu and Yanhong Xu and Yanping Huang and Yaohui Li and Yi Zheng and Yuchen Zhu and Yunxian Ma and Ying Tang and Yukun Zha and Yuting Yan and Z. Z. Ren and Zehui Ren and Zhangli Sha and Zhe Fu and Zhean Xu and Zhenda Xie and Zhengyan Zhang and Zhewen Hao and Zhicheng Ma and Zhigang Yan and Zhiyu Wu and Zihui Gu and Zijia Zhu and Zijun Liu and Zilin Li and Ziwei Xie and Ziyang Song and Zizheng Pan and Zhen Huang and Zhipeng Xu and Zhongyu Zhang and Zhen Zhang},
      year={2025},
      eprint={2501.12948},
      archivePrefix={arXiv},
      primaryClass={cs.CL},
      url={https://arxiv.org/abs/2501.12948}, 
}

@misc{gemmateam2025gemma3technicalreport,
      title={Gemma 3 Technical Report}, 
      author={Gemma Team and Aishwarya Kamath and Johan Ferret and Shreya Pathak and Nino Vieillard and Ramona Merhej and Sarah Perrin and Tatiana Matejovicova and Alexandre Ramé and Morgane Rivière and Louis Rouillard and Thomas Mesnard and Geoffrey Cideron and Jean-bastien Grill and Sabela Ramos and Edouard Yvinec and Michelle Casbon and Etienne Pot and Ivo Penchev and Gaël Liu and Francesco Visin and Kathleen Kenealy and Lucas Beyer and Xiaohai Zhai and Anton Tsitsulin and Robert Busa-Fekete and Alex Feng and Noveen Sachdeva and Benjamin Coleman and Yi Gao and Basil Mustafa and Iain Barr and Emilio Parisotto and David Tian and Matan Eyal and Colin Cherry and Jan-Thorsten Peter and Danila Sinopalnikov and Surya Bhupatiraju and Rishabh Agarwal and Mehran Kazemi and Dan Malkin and Ravin Kumar and David Vilar and Idan Brusilovsky and Jiaming Luo and Andreas Steiner and Abe Friesen and Abhanshu Sharma and Abheesht Sharma and Adi Mayrav Gilady and Adrian Goedeckemeyer and Alaa Saade and Alex Feng and Alexander Kolesnikov and Alexei Bendebury and Alvin Abdagic and Amit Vadi and András György and André Susano Pinto and Anil Das and Ankur Bapna and Antoine Miech and Antoine Yang and Antonia Paterson and Ashish Shenoy and Ayan Chakrabarti and Bilal Piot and Bo Wu and Bobak Shahriari and Bryce Petrini and Charlie Chen and Charline Le Lan and Christopher A. Choquette-Choo and CJ Carey and Cormac Brick and Daniel Deutsch and Danielle Eisenbud and Dee Cattle and Derek Cheng and Dimitris Paparas and Divyashree Shivakumar Sreepathihalli and Doug Reid and Dustin Tran and Dustin Zelle and Eric Noland and Erwin Huizenga and Eugene Kharitonov and Frederick Liu and Gagik Amirkhanyan and Glenn Cameron and Hadi Hashemi and Hanna Klimczak-Plucińska and Harman Singh and Harsh Mehta and Harshal Tushar Lehri and Hussein Hazimeh and Ian Ballantyne and Idan Szpektor and Ivan Nardini and Jean Pouget-Abadie and Jetha Chan and Joe Stanton and John Wieting and Jonathan Lai and Jordi Orbay and Joseph Fernandez and Josh Newlan and Ju-yeong Ji and Jyotinder Singh and Kat Black and Kathy Yu and Kevin Hui and Kiran Vodrahalli and Klaus Greff and Linhai Qiu and Marcella Valentine and Marina Coelho and Marvin Ritter and Matt Hoffman and Matthew Watson and Mayank Chaturvedi and Michael Moynihan and Min Ma and Nabila Babar and Natasha Noy and Nathan Byrd and Nick Roy and Nikola Momchev and Nilay Chauhan and Noveen Sachdeva and Oskar Bunyan and Pankil Botarda and Paul Caron and Paul Kishan Rubenstein and Phil Culliton and Philipp Schmid and Pier Giuseppe Sessa and Pingmei Xu and Piotr Stanczyk and Pouya Tafti and Rakesh Shivanna and Renjie Wu and Renke Pan and Reza Rokni and Rob Willoughby and Rohith Vallu and Ryan Mullins and Sammy Jerome and Sara Smoot and Sertan Girgin and Shariq Iqbal and Shashir Reddy and Shruti Sheth and Siim Põder and Sijal Bhatnagar and Sindhu Raghuram Panyam and Sivan Eiger and Susan Zhang and Tianqi Liu and Trevor Yacovone and Tyler Liechty and Uday Kalra and Utku Evci and Vedant Misra and Vincent Roseberry and Vlad Feinberg and Vlad Kolesnikov and Woohyun Han and Woosuk Kwon and Xi Chen and Yinlam Chow and Yuvein Zhu and Zichuan Wei and Zoltan Egyed and Victor Cotruta and Minh Giang and Phoebe Kirk and Anand Rao and Kat Black and Nabila Babar and Jessica Lo and Erica Moreira and Luiz Gustavo Martins and Omar Sanseviero and Lucas Gonzalez and Zach Gleicher and Tris Warkentin and Vahab Mirrokni and Evan Senter and Eli Collins and Joelle Barral and Zoubin Ghahramani and Raia Hadsell and Yossi Matias and D. Sculley and Slav Petrov and Noah Fiedel and Noam Shazeer and Oriol Vinyals and Jeff Dean and Demis Hassabis and Koray Kavukcuoglu and Clement Farabet and Elena Buchatskaya and Jean-Baptiste Alayrac and Rohan Anil and Dmitry and Lepikhin and Sebastian Borgeaud and Olivier Bachem and Armand Joulin and Alek Andreev and Cassidy Hardin and Robert Dadashi and Léonard Hussenot},
      year={2025},
      eprint={2503.19786},
      archivePrefix={arXiv},
      primaryClass={cs.CL},
      url={https://arxiv.org/abs/2503.19786}, 
}

@misc{grattafiori2024llama3herdmodels,
      title={The Llama 3 Herd of Models}, 
      author={Aaron Grattafiori and Abhimanyu Dubey and Abhinav Jauhri and Abhinav Pandey and Abhishek Kadian and Ahmad Al-Dahle and Aiesha Letman and Akhil Mathur and Alan Schelten and Alex Vaughan and Amy Yang and Angela Fan and Anirudh Goyal and Anthony Hartshorn and Aobo Yang and Archi Mitra and Archie Sravankumar and Artem Korenev and Arthur Hinsvark and Arun Rao and Aston Zhang and Aurelien Rodriguez and Austen Gregerson and Ava Spataru and Baptiste Roziere and Bethany Biron and Binh Tang and Bobbie Chern and Charlotte Caucheteux and Chaya Nayak and Chloe Bi and Chris Marra and Chris McConnell and Christian Keller and Christophe Touret and Chunyang Wu and Corinne Wong and Cristian Canton Ferrer and Cyrus Nikolaidis and Damien Allonsius and Daniel Song and Danielle Pintz and Danny Livshits and Danny Wyatt and David Esiobu and Dhruv Choudhary and Dhruv Mahajan and Diego Garcia-Olano and Diego Perino and Dieuwke Hupkes and Egor Lakomkin and Ehab AlBadawy and Elina Lobanova and Emily Dinan and Eric Michael Smith and Filip Radenovic and Francisco Guzmán and Frank Zhang and Gabriel Synnaeve and Gabrielle Lee and Georgia Lewis Anderson and Govind Thattai and Graeme Nail and Gregoire Mialon and Guan Pang and Guillem Cucurell and Hailey Nguyen and Hannah Korevaar and Hu Xu and Hugo Touvron and Iliyan Zarov and Imanol Arrieta Ibarra and Isabel Kloumann and Ishan Misra and Ivan Evtimov and Jack Zhang and Jade Copet and Jaewon Lee and Jan Geffert and Jana Vranes and Jason Park and Jay Mahadeokar and Jeet Shah and Jelmer van der Linde and Jennifer Billock and Jenny Hong and Jenya Lee and Jeremy Fu and Jianfeng Chi and Jianyu Huang and Jiawen Liu and Jie Wang and Jiecao Yu and Joanna Bitton and Joe Spisak and Jongsoo Park and Joseph Rocca and Joshua Johnstun and Joshua Saxe and Junteng Jia and Kalyan Vasuden Alwala and Karthik Prasad and Kartikeya Upasani and Kate Plawiak and Ke Li and Kenneth Heafield and Kevin Stone and Khalid El-Arini and Krithika Iyer and Kshitiz Malik and Kuenley Chiu and Kunal Bhalla and Kushal Lakhotia and Lauren Rantala-Yeary and Laurens van der Maaten and Lawrence Chen and Liang Tan and Liz Jenkins and Louis Martin and Lovish Madaan and Lubo Malo and Lukas Blecher and Lukas Landzaat and Luke de Oliveira and Madeline Muzzi and Mahesh Pasupuleti and Mannat Singh and Manohar Paluri and Marcin Kardas and Maria Tsimpoukelli and Mathew Oldham and Mathieu Rita and Maya Pavlova and Melanie Kambadur and Mike Lewis and Min Si and Mitesh Kumar Singh and Mona Hassan and Naman Goyal and Narjes Torabi and Nikolay Bashlykov and Nikolay Bogoychev and Niladri Chatterji and Ning Zhang and Olivier Duchenne and Onur Çelebi and Patrick Alrassy and Pengchuan Zhang and Pengwei Li and Petar Vasic and Peter Weng and Prajjwal Bhargava and Pratik Dubal and Praveen Krishnan and Punit Singh Koura and Puxin Xu and Qing He and Qingxiao Dong and Ragavan Srinivasan and Raj Ganapathy and Ramon Calderer and Ricardo Silveira Cabral and Robert Stojnic and Roberta Raileanu and Rohan Maheswari and Rohit Girdhar and Rohit Patel and Romain Sauvestre and Ronnie Polidoro and Roshan Sumbaly and Ross Taylor and Ruan Silva and Rui Hou and Rui Wang and Saghar Hosseini and Sahana Chennabasappa and Sanjay Singh and Sean Bell and Seohyun Sonia Kim and Sergey Edunov and Shaoliang Nie and Sharan Narang and Sharath Raparthy and Sheng Shen and Shengye Wan and Shruti Bhosale and Shun Zhang and Simon Vandenhende and Soumya Batra and Spencer Whitman and Sten Sootla and Stephane Collot and Suchin Gururangan and Sydney Borodinsky and Tamar Herman and Tara Fowler and Tarek Sheasha and Thomas Georgiou and Thomas Scialom and Tobias Speckbacher and Todor Mihaylov and Tong Xiao and Ujjwal Karn and Vedanuj Goswami and Vibhor Gupta and Vignesh Ramanathan and Viktor Kerkez and Vincent Gonguet and Virginie Do and Vish Vogeti and Vítor Albiero and Vladan Petrovic and Weiwei Chu and Wenhan Xiong and Wenyin Fu and Whitney Meers and Xavier Martinet and Xiaodong Wang and Xiaofang Wang and Xiaoqing Ellen Tan and Xide Xia and Xinfeng Xie and Xuchao Jia and Xuewei Wang and Yaelle Goldschlag and Yashesh Gaur and Yasmine Babaei and Yi Wen and Yiwen Song and Yuchen Zhang and Yue Li and Yuning Mao and Zacharie Delpierre Coudert and Zheng Yan and Zhengxing Chen and Zoe Papakipos and Aaditya Singh and Aayushi Srivastava and Abha Jain and Adam Kelsey and Adam Shajnfeld and Adithya Gangidi and Adolfo Victoria and Ahuva Goldstand and Ajay Menon and Ajay Sharma and Alex Boesenberg and Alexei Baevski and Allie Feinstein and Amanda Kallet and Amit Sangani and Amos Teo and Anam Yunus and Andrei Lupu and Andres Alvarado and Andrew Caples and Andrew Gu and Andrew Ho and Andrew Poulton and Andrew Ryan and Ankit Ramchandani and Annie Dong and Annie Franco and Anuj Goyal and Aparajita Saraf and Arkabandhu Chowdhury and Ashley Gabriel and Ashwin Bharambe and Assaf Eisenman and Azadeh Yazdan and Beau James and Ben Maurer and Benjamin Leonhardi and Bernie Huang and Beth Loyd and Beto De Paola and Bhargavi Paranjape and Bing Liu and Bo Wu and Boyu Ni and Braden Hancock and Bram Wasti and Brandon Spence and Brani Stojkovic and Brian Gamido and Britt Montalvo and Carl Parker and Carly Burton and Catalina Mejia and Ce Liu and Changhan Wang and Changkyu Kim and Chao Zhou and Chester Hu and Ching-Hsiang Chu and Chris Cai and Chris Tindal and Christoph Feichtenhofer and Cynthia Gao and Damon Civin and Dana Beaty and Daniel Kreymer and Daniel Li and David Adkins and David Xu and Davide Testuggine and Delia David and Devi Parikh and Diana Liskovich and Didem Foss and Dingkang Wang and Duc Le and Dustin Holland and Edward Dowling and Eissa Jamil and Elaine Montgomery and Eleonora Presani and Emily Hahn and Emily Wood and Eric-Tuan Le and Erik Brinkman and Esteban Arcaute and Evan Dunbar and Evan Smothers and Fei Sun and Felix Kreuk and Feng Tian and Filippos Kokkinos and Firat Ozgenel and Francesco Caggioni and Frank Kanayet and Frank Seide and Gabriela Medina Florez and Gabriella Schwarz and Gada Badeer and Georgia Swee and Gil Halpern and Grant Herman and Grigory Sizov and Guangyi and Zhang and Guna Lakshminarayanan and Hakan Inan and Hamid Shojanazeri and Han Zou and Hannah Wang and Hanwen Zha and Haroun Habeeb and Harrison Rudolph and Helen Suk and Henry Aspegren and Hunter Goldman and Hongyuan Zhan and Ibrahim Damlaj and Igor Molybog and Igor Tufanov and Ilias Leontiadis and Irina-Elena Veliche and Itai Gat and Jake Weissman and James Geboski and James Kohli and Janice Lam and Japhet Asher and Jean-Baptiste Gaya and Jeff Marcus and Jeff Tang and Jennifer Chan and Jenny Zhen and Jeremy Reizenstein and Jeremy Teboul and Jessica Zhong and Jian Jin and Jingyi Yang and Joe Cummings and Jon Carvill and Jon Shepard and Jonathan McPhie and Jonathan Torres and Josh Ginsburg and Junjie Wang and Kai Wu and Kam Hou U and Karan Saxena and Kartikay Khandelwal and Katayoun Zand and Kathy Matosich and Kaushik Veeraraghavan and Kelly Michelena and Keqian Li and Kiran Jagadeesh and Kun Huang and Kunal Chawla and Kyle Huang and Lailin Chen and Lakshya Garg and Lavender A and Leandro Silva and Lee Bell and Lei Zhang and Liangpeng Guo and Licheng Yu and Liron Moshkovich and Luca Wehrstedt and Madian Khabsa and Manav Avalani and Manish Bhatt and Martynas Mankus and Matan Hasson and Matthew Lennie and Matthias Reso and Maxim Groshev and Maxim Naumov and Maya Lathi and Meghan Keneally and Miao Liu and Michael L. Seltzer and Michal Valko and Michelle Restrepo and Mihir Patel and Mik Vyatskov and Mikayel Samvelyan and Mike Clark and Mike Macey and Mike Wang and Miquel Jubert Hermoso and Mo Metanat and Mohammad Rastegari and Munish Bansal and Nandhini Santhanam and Natascha Parks and Natasha White and Navyata Bawa and Nayan Singhal and Nick Egebo and Nicolas Usunier and Nikhil Mehta and Nikolay Pavlovich Laptev and Ning Dong and Norman Cheng and Oleg Chernoguz and Olivia Hart and Omkar Salpekar and Ozlem Kalinli and Parkin Kent and Parth Parekh and Paul Saab and Pavan Balaji and Pedro Rittner and Philip Bontrager and Pierre Roux and Piotr Dollar and Polina Zvyagina and Prashant Ratanchandani and Pritish Yuvraj and Qian Liang and Rachad Alao and Rachel Rodriguez and Rafi Ayub and Raghotham Murthy and Raghu Nayani and Rahul Mitra and Rangaprabhu Parthasarathy and Raymond Li and Rebekkah Hogan and Robin Battey and Rocky Wang and Russ Howes and Ruty Rinott and Sachin Mehta and Sachin Siby and Sai Jayesh Bondu and Samyak Datta and Sara Chugh and Sara Hunt and Sargun Dhillon and Sasha Sidorov and Satadru Pan and Saurabh Mahajan and Saurabh Verma and Seiji Yamamoto and Sharadh Ramaswamy and Shaun Lindsay and Shaun Lindsay and Sheng Feng and Shenghao Lin and Shengxin Cindy Zha and Shishir Patil and Shiva Shankar and Shuqiang Zhang and Shuqiang Zhang and Sinong Wang and Sneha Agarwal and Soji Sajuyigbe and Soumith Chintala and Stephanie Max and Stephen Chen and Steve Kehoe and Steve Satterfield and Sudarshan Govindaprasad and Sumit Gupta and Summer Deng and Sungmin Cho and Sunny Virk and Suraj Subramanian and Sy Choudhury and Sydney Goldman and Tal Remez and Tamar Glaser and Tamara Best and Thilo Koehler and Thomas Robinson and Tianhe Li and Tianjun Zhang and Tim Matthews and Timothy Chou and Tzook Shaked and Varun Vontimitta and Victoria Ajayi and Victoria Montanez and Vijai Mohan and Vinay Satish Kumar and Vishal Mangla and Vlad Ionescu and Vlad Poenaru and Vlad Tiberiu Mihailescu and Vladimir Ivanov and Wei Li and Wenchen Wang and Wenwen Jiang and Wes Bouaziz and Will Constable and Xiaocheng Tang and Xiaojian Wu and Xiaolan Wang and Xilun Wu and Xinbo Gao and Yaniv Kleinman and Yanjun Chen and Ye Hu and Ye Jia and Ye Qi and Yenda Li and Yilin Zhang and Ying Zhang and Yossi Adi and Youngjin Nam and Yu and Wang and Yu Zhao and Yuchen Hao and Yundi Qian and Yunlu Li and Yuzi He and Zach Rait and Zachary DeVito and Zef Rosnbrick and Zhaoduo Wen and Zhenyu Yang and Zhiwei Zhao and Zhiyu Ma},
      year={2024},
      eprint={2407.21783},
      archivePrefix={arXiv},
      primaryClass={cs.AI},
      url={https://arxiv.org/abs/2407.21783}, 
}

@misc{yang2025qwen3technicalreport,
      title={Qwen3 Technical Report}, 
      author={An Yang and Anfeng Li and Baosong Yang and Beichen Zhang and Binyuan Hui and Bo Zheng and Bowen Yu and Chang Gao and Chengen Huang and Chenxu Lv and Chujie Zheng and Dayiheng Liu and Fan Zhou and Fei Huang and Feng Hu and Hao Ge and Haoran Wei and Huan Lin and Jialong Tang and Jian Yang and Jianhong Tu and Jianwei Zhang and Jianxin Yang and Jiaxi Yang and Jing Zhou and Jingren Zhou and Junyang Lin and Kai Dang and Keqin Bao and Kexin Yang and Le Yu and Lianghao Deng and Mei Li and Mingfeng Xue and Mingze Li and Pei Zhang and Peng Wang and Qin Zhu and Rui Men and Ruize Gao and Shixuan Liu and Shuang Luo and Tianhao Li and Tianyi Tang and Wenbiao Yin and Xingzhang Ren and Xinyu Wang and Xinyu Zhang and Xuancheng Ren and Yang Fan and Yang Su and Yichang Zhang and Yinger Zhang and Yu Wan and Yuqiong Liu and Zekun Wang and Zeyu Cui and Zhenru Zhang and Zhipeng Zhou and Zihan Qiu},
      year={2025},
      eprint={2505.09388},
      archivePrefix={arXiv},
      primaryClass={cs.CL},
      url={https://arxiv.org/abs/2505.09388}, 
}

@misc{bowman2015largeannotatedcorpuslearning,
      title={A large annotated corpus for learning natural language inference}, 
      author={Samuel R. Bowman and Gabor Angeli and Christopher Potts and Christopher D. Manning},
      year={2015},
      eprint={1508.05326},
      archivePrefix={arXiv},
      primaryClass={cs.CL},
      url={https://arxiv.org/abs/1508.05326}, 
}

@inproceedings{wu-etal-2023-hence,
    title = "Hence, Socrates is mortal: A Benchmark for Natural Language Syllogistic Reasoning",
    author = "Wu, Yongkang  and
      Han, Meng  and
      Zhu, Yutao  and
      Li, Lei  and
      Zhang, Xinyu  and
      Lai, Ruofei  and
      Li, Xiaoguang  and
      Ren, Yuanhang  and
      Dou, Zhicheng  and
      Cao, Zhao",
    editor = "Rogers, Anna  and
      Boyd-Graber, Jordan  and
      Okazaki, Naoaki",
    booktitle = "Findings of the Association for Computational Linguistics: ACL 2023",
    month = jul,
    year = "2023",
    address = "Toronto, Canada",
    publisher = "Association for Computational Linguistics",
    url = "https://aclanthology.org/2023.findings-acl.148/",
    doi = "10.18653/v1/2023.findings-acl.148",
    pages = "2347--2367"
}

@misc{nguyen2025bisreasoning,
      title={BIS Reasoning 1.0: The First Large-Scale Japanese Benchmark for Belief-Inconsistent Syllogistic Reasoning}, 
      author={Ha-Thanh Nguyen and Chaoran Liu and Qianying Liu and Hideyuki Tachibana and Su Myat Noe and Yusuke Miyao and Koichi Takeda and Sadao Kurohashi},
      year={2025},
      eprint={2506.06955},
      archivePrefix={arXiv},
      primaryClass={cs.CL},
      url={https://arxiv.org/abs/2506.06955}, 
}

@misc{han2024folio,
      title={FOLIO: Natural Language Reasoning with First-Order Logic}, 
      author={Simeng Han and Hailey Schoelkopf and Yilun Zhao and Zhenting Qi and Martin Riddell and Wenfei Zhou and James Coady and David Peng and Yujie Qiao and Luke Benson and Lucy Sun and Alex Wardle-Solano and Hannah Szabo and Ekaterina Zubova and Matthew Burtell and Jonathan Fan and Yixin Liu and Brian Wong and Malcolm Sailor and Ansong Ni and Linyong Nan and Jungo Kasai and Tao Yu and Rui Zhang and Alexander R. Fabbri and Wojciech Kryscinski and Semih Yavuz and Ye Liu and Xi Victoria Lin and Shafiq Joty and Yingbo Zhou and Caiming Xiong and Rex Ying and Arman Cohan and Dragomir Radev},
      year={2024},
      eprint={2209.00840},
      archivePrefix={arXiv},
      primaryClass={cs.CL},
      url={https://arxiv.org/abs/2209.00840}, 
}

@misc{ando2023NeuBAROCO,
      title={Evaluating Large Language Models with NeuBAROCO: Syllogistic Reasoning Ability and Human-like Biases}, 
      author={Risako Ando and Takanobu Morishita and Hirohiko Abe and Koji Mineshima and Mitsuhiro Okada},
      year={2023},
      eprint={2306.12567},
      archivePrefix={arXiv},
      primaryClass={cs.CL},
      url={https://arxiv.org/abs/2306.12567}, 
}

@article{SHIKISHIMA2009256,
title = {Is g an entity? A Japanese twin study using syllogisms and intelligence tests},
journal = {Intelligence},
volume = {37},
number = {3},
pages = {256-267},
year = {2009},
issn = {0160-2896},
doi = {https://doi.org/10.1016/j.intell.2008.10.010},
url = {https://www.sciencedirect.com/science/article/pii/S0160289608001591},
author = {Chizuru Shikishima and Kai Hiraishi and Shinji Yamagata and Yutaro Sugimoto and Ryo Takemura and Koken Ozaki and Mitsuhiro Okada and Tatsushi Toda and Juko Ando}
}

@misc{wei2023cmath,
      title={CMATH: Can Your Language Model Pass Chinese Elementary School Math Test?}, 
      author={Tianwen Wei and Jian Luan and Wei Liu and Shuang Dong and Bin Wang},
      year={2023},
      eprint={2306.16636},
      archivePrefix={arXiv},
      primaryClass={cs.CL},
      url={https://arxiv.org/abs/2306.16636}, 
}

@misc{bisk2019piqa,
      title={PIQA: Reasoning about Physical Commonsense in Natural Language}, 
      author={Yonatan Bisk and Rowan Zellers and Ronan Le Bras and Jianfeng Gao and Yejin Choi},
      year={2019},
      eprint={1911.11641},
      archivePrefix={arXiv},
      primaryClass={cs.CL},
      url={https://arxiv.org/abs/1911.11641}, 
}

@misc{bang2023multitask,
      title={A Multitask, Multilingual, Multimodal Evaluation of ChatGPT on Reasoning, Hallucination, and Interactivity}, 
      author={Yejin Bang and Samuel Cahyawijaya and Nayeon Lee and Wenliang Dai and Dan Su and Bryan Wilie and Holy Lovenia and Ziwei Ji and Tiezheng Yu and Willy Chung and Quyet V. Do and Yan Xu and Pascale Fung},
      year={2023},
      eprint={2302.04023},
      archivePrefix={arXiv},
      primaryClass={cs.CL},
      url={https://arxiv.org/abs/2302.04023}, 
}

@misc{phan2025humanitysexam,
      title={Humanity's Last Exam}, 
      author={Long Phan and Alice Gatti and Ziwen Han and Nathaniel Li and Josephina Hu and Hugh Zhang and Chen Bo Calvin Zhang and Mohamed Shaaban and John Ling and Sean Shi and Michael Choi and Anish Agrawal and Arnav Chopra and Adam Khoja and Ryan Kim and Richard Ren and Jason Hausenloy and Oliver Zhang and Mantas Mazeika and Dmitry Dodonov and Tung Nguyen and Jaeho Lee and Daron Anderson and Mikhail Doroshenko and Alun Cennyth Stokes and Mobeen Mahmood and Oleksandr Pokutnyi and Oleg Iskra and Jessica P. Wang and John-Clark Levin and Mstyslav Kazakov and Fiona Feng and Steven Y. Feng and Haoran Zhao and Michael Yu and Varun Gangal and Chelsea Zou and Zihan Wang and Serguei Popov and Robert Gerbicz and Geoff Galgon and Johannes Schmitt and Will Yeadon and Yongki Lee and Scott Sauers and Alvaro Sanchez and Fabian Giska and Marc Roth and Søren Riis and Saiteja Utpala and Noah Burns and Gashaw M. Goshu and Mohinder Maheshbhai Naiya and Chidozie Agu and Zachary Giboney and Antrell Cheatom and Francesco Fournier-Facio and Sarah-Jane Crowson and Lennart Finke and Zerui Cheng and Jennifer Zampese and Ryan G. Hoerr and Mark Nandor and Hyunwoo Park and Tim Gehrunger and Jiaqi Cai and Ben McCarty and Alexis C Garretson and Edwin Taylor and Damien Sileo and Qiuyu Ren and Usman Qazi and Lianghui Li and Jungbae Nam and John B. Wydallis and Pavel Arkhipov and Jack Wei Lun Shi and Aras Bacho and Chris G. Willcocks and Hangrui Cao and Sumeet Motwani and Emily de Oliveira Santos and Johannes Veith and Edward Vendrow and Doru Cojoc and Kengo Zenitani and Joshua Robinson and Longke Tang and Yuqi Li and Joshua Vendrow and Natanael Wildner Fraga and Vladyslav Kuchkin and Andrey Pupasov Maksimov and Pierre Marion and Denis Efremov and Jayson Lynch and Kaiqu Liang and Aleksandar Mikov and Andrew Gritsevskiy and Julien Guillod and Gözdenur Demir and Dakotah Martinez and Ben Pageler and Kevin Zhou and Saeed Soori and Ori Press and Henry Tang and Paolo Rissone and Sean R. Green and Lina Brüssel and Moon Twayana and Aymeric Dieuleveut and Joseph Marvin Imperial and Ameya Prabhu and Jinzhou Yang and Nick Crispino and Arun Rao and Dimitri Zvonkine and Gabriel Loiseau and Mikhail Kalinin and Marco Lukas and Ciprian Manolescu and Nate Stambaugh and Subrata Mishra and Tad Hogg and Carlo Bosio and Brian P Coppola and Julian Salazar and Jaehyeok Jin and Rafael Sayous and Stefan Ivanov and Philippe Schwaller and Shaipranesh Senthilkuma and Andres M Bran and Andres Algaba and Kelsey Van den Houte and Lynn Van Der Sypt and Brecht Verbeken and David Noever and Alexei Kopylov and Benjamin Myklebust and Bikun Li and Lisa Schut and Evgenii Zheltonozhskii and Qiaochu Yuan and Derek Lim and Richard Stanley and Tong Yang and John Maar and Julian Wykowski and Martí Oller and Anmol Sahu and Cesare Giulio Ardito and Yuzheng Hu and Ariel Ghislain Kemogne Kamdoum and Alvin Jin and Tobias Garcia Vilchis and Yuexuan Zu and Martin Lackner and James Koppel and Gongbo Sun and Daniil S. Antonenko and Steffi Chern and Bingchen Zhao and Pierrot Arsene and Joseph M Cavanagh and Daofeng Li and Jiawei Shen and Donato Crisostomi and Wenjin Zhang and Ali Dehghan and Sergey Ivanov and David Perrella and Nurdin Kaparov and Allen Zang and Ilia Sucholutsky and Arina Kharlamova and Daniil Orel and Vladislav Poritski and Shalev Ben-David and Zachary Berger and Parker Whitfill and Michael Foster and Daniel Munro and Linh Ho and Shankar Sivarajan and Dan Bar Hava and Aleksey Kuchkin and David Holmes and Alexandra Rodriguez-Romero and Frank Sommerhage and Anji Zhang and Richard Moat and Keith Schneider and Zakayo Kazibwe and Don Clarke and Dae Hyun Kim and Felipe Meneguitti Dias and Sara Fish and Veit Elser and Tobias Kreiman and Victor Efren Guadarrama Vilchis and Immo Klose and Ujjwala Anantheswaran and Adam Zweiger and Kaivalya Rawal and Jeffery Li and Jeremy Nguyen and Nicolas Daans and Haline Heidinger and Maksim Radionov and Václav Rozhoň and Vincent Ginis and Christian Stump and Niv Cohen and Rafał Poświata and Josef Tkadlec and Alan Goldfarb and Chenguang Wang and Piotr Padlewski and Stanislaw Barzowski and Kyle Montgomery and Ryan Stendall and Jamie Tucker-Foltz and Jack Stade and T. Ryan Rogers and Tom Goertzen and Declan Grabb and Abhishek Shukla and Alan Givré and John Arnold Ambay and Archan Sen and Muhammad Fayez Aziz and Mark H Inlow and Hao He and Ling Zhang and Younesse Kaddar and Ivar Ängquist and Yanxu Chen and Harrison K Wang and Kalyan Ramakrishnan and Elliott Thornley and Antonio Terpin and Hailey Schoelkopf and Eric Zheng and Avishy Carmi and Ethan D. L. Brown and Kelin Zhu and Max Bartolo and Richard Wheeler and Martin Stehberger and Peter Bradshaw and JP Heimonen and Kaustubh Sridhar and Ido Akov and Jennifer Sandlin and Yury Makarychev and Joanna Tam and Hieu Hoang and David M. Cunningham and Vladimir Goryachev and Demosthenes Patramanis and Michael Krause and Andrew Redenti and David Aldous and Jesyin Lai and Shannon Coleman and Jiangnan Xu and Sangwon Lee and Ilias Magoulas and Sandy Zhao and Ning Tang and Michael K. Cohen and Orr Paradise and Jan Hendrik Kirchner and Maksym Ovchynnikov and Jason O. Matos and Adithya Shenoy and Michael Wang and Yuzhou Nie and Anna Sztyber-Betley and Paolo Faraboschi and Robin Riblet and Jonathan Crozier and Shiv Halasyamani and Shreyas Verma and Prashant Joshi and Eli Meril and Ziqiao Ma and Jérémy Andréoletti and Raghav Singhal and Jacob Platnick and Volodymyr Nevirkovets and Luke Basler and Alexander Ivanov and Seri Khoury and Nils Gustafsson and Marco Piccardo and Hamid Mostaghimi and Qijia Chen and Virendra Singh and Tran Quoc Khánh and Paul Rosu and Hannah Szlyk and Zachary Brown and Himanshu Narayan and Aline Menezes and Jonathan Roberts and William Alley and Kunyang Sun and Arkil Patel and Max Lamparth and Anka Reuel and Linwei Xin and Hanmeng Xu and Jacob Loader and Freddie Martin and Zixuan Wang and Andrea Achilleos and Thomas Preu and Tomek Korbak and Ida Bosio and Fereshteh Kazemi and Ziye Chen and Biró Bálint and Eve J. Y. Lo and Jiaqi Wang and Maria Inês S. Nunes and Jeremiah Milbauer and M Saiful Bari and Zihao Wang and Behzad Ansarinejad and Yewen Sun and Stephane Durand and Hossam Elgnainy and Guillaume Douville and Daniel Tordera and George Balabanian and Hew Wolff and Lynna Kvistad and Hsiaoyun Milliron and Ahmad Sakor and Murat Eron and Andrew Favre D. O. and Shailesh Shah and Xiaoxiang Zhou and Firuz Kamalov and Sherwin Abdoli and Tim Santens and Shaul Barkan and Allison Tee and Robin Zhang and Alessandro Tomasiello and G. Bruno De Luca and Shi-Zhuo Looi and Vinh-Kha Le and Noam Kolt and Jiayi Pan and Emma Rodman and Jacob Drori and Carl J Fossum and Niklas Muennighoff and Milind Jagota and Ronak Pradeep and Honglu Fan and Jonathan Eicher and Michael Chen and Kushal Thaman and William Merrill and Moritz Firsching and Carter Harris and Stefan Ciobâcă and Jason Gross and Rohan Pandey and Ilya Gusev and Adam Jones and Shashank Agnihotri and Pavel Zhelnov and Mohammadreza Mofayezi and Alexander Piperski and David K. Zhang and Kostiantyn Dobarskyi and Roman Leventov and Ignat Soroko and Joshua Duersch and Vage Taamazyan and Andrew Ho and Wenjie Ma and William Held and Ruicheng Xian and Armel Randy Zebaze and Mohanad Mohamed and Julian Noah Leser and Michelle X Yuan and Laila Yacar and Johannes Lengler and Katarzyna Olszewska and Claudio Di Fratta and Edson Oliveira and Joseph W. Jackson and Andy Zou and Muthu Chidambaram and Timothy Manik and Hector Haffenden and Dashiell Stander and Ali Dasouqi and Alexander Shen and Bita Golshani and David Stap and Egor Kretov and Mikalai Uzhou and Alina Borisovna Zhidkovskaya and Nick Winter and Miguel Orbegozo Rodriguez and Robert Lauff and Dustin Wehr and Colin Tang and Zaki Hossain and Shaun Phillips and Fortuna Samuele and Fredrik Ekström and Angela Hammon and Oam Patel and Faraz Farhidi and George Medley and Forough Mohammadzadeh and Madellene Peñaflor and Haile Kassahun and Alena Friedrich and Rayner Hernandez Perez and Daniel Pyda and Taom Sakal and Omkar Dhamane and Ali Khajegili Mirabadi and Eric Hallman and Kenchi Okutsu and Mike Battaglia and Mohammad Maghsoudimehrabani and Alon Amit and Dave Hulbert and Roberto Pereira and Simon Weber and Handoko and Anton Peristyy and Stephen Malina and Mustafa Mehkary and Rami Aly and Frank Reidegeld and Anna-Katharina Dick and Cary Friday and Mukhwinder Singh and Hassan Shapourian and Wanyoung Kim and Mariana Costa and Hubeyb Gurdogan and Harsh Kumar and Chiara Ceconello and Chao Zhuang and Haon Park and Micah Carroll and Andrew R. Tawfeek and Stefan Steinerberger and Daattavya Aggarwal and Michael Kirchhof and Linjie Dai and Evan Kim and Johan Ferret and Jainam Shah and Yuzhou Wang and Minghao Yan and Krzysztof Burdzy and Lixin Zhang and Antonio Franca and Diana T. Pham and Kang Yong Loh and Joshua Robinson and Abram Jackson and Paolo Giordano and Philipp Petersen and Adrian Cosma and Jesus Colino and Colin White and Jacob Votava and Vladimir Vinnikov and Ethan Delaney and Petr Spelda and Vit Stritecky and Syed M. Shahid and Jean-Christophe Mourrat and Lavr Vetoshkin and Koen Sponselee and Renas Bacho and Zheng-Xin Yong and Florencia de la Rosa and Nathan Cho and Xiuyu Li and Guillaume Malod and Orion Weller and Guglielmo Albani and Leon Lang and Julien Laurendeau and Dmitry Kazakov and Fatimah Adesanya and Julien Portier and Lawrence Hollom and Victor Souza and Yuchen Anna Zhou and Julien Degorre and Yiğit Yalın and Gbenga Daniel Obikoya and Rai and Filippo Bigi and M. C. Boscá and Oleg Shumar and Kaniuar Bacho and Gabriel Recchia and Mara Popescu and Nikita Shulga and Ngefor Mildred Tanwie and Thomas C. H. Lux and Ben Rank and Colin Ni and Matthew Brooks and Alesia Yakimchyk and Huanxu and Liu and Stefano Cavalleri and Olle Häggström and Emil Verkama and Joshua Newbould and Hans Gundlach and Leonor Brito-Santana and Brian Amaro and Vivek Vajipey and Rynaa Grover and Ting Wang and Yosi Kratish and Wen-Ding Li and Sivakanth Gopi and Andrea Caciolai and Christian Schroeder de Witt and Pablo Hernández-Cámara and Emanuele Rodolà and Jules Robins and Dominic Williamson and Vincent Cheng and Brad Raynor and Hao Qi and Ben Segev and Jingxuan Fan and Sarah Martinson and Erik Y. Wang and Kaylie Hausknecht and Michael P. Brenner and Mao Mao and Christoph Demian and Peyman Kassani and Xinyu Zhang and David Avagian and Eshawn Jessica Scipio and Alon Ragoler and Justin Tan and Blake Sims and Rebeka Plecnik and Aaron Kirtland and Omer Faruk Bodur and D. P. Shinde and Yan Carlos Leyva Labrador and Zahra Adoul and Mohamed Zekry and Ali Karakoc and Tania C. B. Santos and Samir Shamseldeen and Loukmane Karim and Anna Liakhovitskaia and Nate Resman and Nicholas Farina and Juan Carlos Gonzalez and Gabe Maayan and Earth Anderson and Rodrigo De Oliveira Pena and Elizabeth Kelley and Hodjat Mariji and Rasoul Pouriamanesh and Wentao Wu and Ross Finocchio and Ismail Alarab and Joshua Cole and Danyelle Ferreira and Bryan Johnson and Mohammad Safdari and Liangti Dai and Siriphan Arthornthurasuk and Isaac C. McAlister and Alejandro José Moyano and Alexey Pronin and Jing Fan and Angel Ramirez-Trinidad and Yana Malysheva and Daphiny Pottmaier and Omid Taheri and Stanley Stepanic and Samuel Perry and Luke Askew and Raúl Adrián Huerta Rodríguez and Ali M. R. Minissi and Ricardo Lorena and Krishnamurthy Iyer and Arshad Anil Fasiludeen and Ronald Clark and Josh Ducey and Matheus Piza and Maja Somrak and Eric Vergo and Juehang Qin and Benjámin Borbás and Eric Chu and Jack Lindsey and Antoine Jallon and I. M. J. McInnis and Evan Chen and Avi Semler and Luk Gloor and Tej Shah and Marc Carauleanu and Pascal Lauer and Tran Đuc Huy and Hossein Shahrtash and Emilien Duc and Lukas Lewark and Assaf Brown and Samuel Albanie and Brian Weber and Warren S. Vaz and Pierre Clavier and Yiyang Fan and Gabriel Poesia Reis e Silva and Long and Lian and Marcus Abramovitch and Xi Jiang and Sandra Mendoza and Murat Islam and Juan Gonzalez and Vasilios Mavroudis and Justin Xu and Pawan Kumar and Laxman Prasad Goswami and Daniel Bugas and Nasser Heydari and Ferenc Jeanplong and Thorben Jansen and Antonella Pinto and Archimedes Apronti and Abdallah Galal and Ng Ze-An and Ankit Singh and Tong Jiang and Joan of Arc Xavier and Kanu Priya Agarwal and Mohammed Berkani and Gang Zhang and Zhehang Du and Benedito Alves de Oliveira Junior and Dmitry Malishev and Nicolas Remy and Taylor D. Hartman and Tim Tarver and Stephen Mensah and Gautier Abou Loume and Wiktor Morak and Farzad Habibi and Sarah Hoback and Will Cai and Javier Gimenez and Roselynn Grace Montecillo and Jakub Łucki and Russell Campbell and Asankhaya Sharma and Khalida Meer and Shreen Gul and Daniel Espinosa Gonzalez and Xavier Alapont and Alex Hoover and Gunjan Chhablani and Freddie Vargus and Arunim Agarwal and Yibo Jiang and Deepakkumar Patil and David Outevsky and Kevin Joseph Scaria and Rajat Maheshwari and Abdelkader Dendane and Priti Shukla and Ashley Cartwright and Sergei Bogdanov and Niels Mündler and Sören Möller and Luca Arnaboldi and Kunvar Thaman and Muhammad Rehan Siddiqi and Prajvi Saxena and Himanshu Gupta and Tony Fruhauff and Glen Sherman and Mátyás Vincze and Siranut Usawasutsakorn and Dylan Ler and Anil Radhakrishnan and Innocent Enyekwe and Sk Md Salauddin and Jiang Muzhen and Aleksandr Maksapetyan and Vivien Rossbach and Chris Harjadi and Mohsen Bahaloohoreh and Claire Sparrow and Jasdeep Sidhu and Sam Ali and Song Bian and John Lai and Eric Singer and Justine Leon Uro and Greg Bateman and Mohamed Sayed and Ahmed Menshawy and Darling Duclosel and Dario Bezzi and Yashaswini Jain and Ashley Aaron and Murat Tiryakioglu and Sheeshram Siddh and Keith Krenek and Imad Ali Shah and Jun Jin and Scott Creighton and Denis Peskoff and Zienab EL-Wasif and Ragavendran P V and Michael Richmond and Joseph McGowan and Tejal Patwardhan and Hao-Yu Sun and Ting Sun and Nikola Zubić and Samuele Sala and Stephen Ebert and Jean Kaddour and Manuel Schottdorf and Dianzhuo Wang and Gerol Petruzella and Alex Meiburg and Tilen Medved and Ali ElSheikh and S Ashwin Hebbar and Lorenzo Vaquero and Xianjun Yang and Jason Poulos and Vilém Zouhar and Sergey Bogdanik and Mingfang Zhang and Jorge Sanz-Ros and David Anugraha and Yinwei Dai and Anh N. Nhu and Xue Wang and Ali Anil Demircali and Zhibai Jia and Yuyin Zhou and Juncheng Wu and Mike He and Nitin Chandok and Aarush Sinha and Gaoxiang Luo and Long Le and Mickaël Noyé and Michał Perełkiewicz and Ioannis Pantidis and Tianbo Qi and Soham Sachin Purohit and Letitia Parcalabescu and Thai-Hoa Nguyen and Genta Indra Winata and Edoardo M. Ponti and Hanchen Li and Kaustubh Dhole and Jongee Park and Dario Abbondanza and Yuanli Wang and Anupam Nayak and Diogo M. Caetano and Antonio A. W. L. Wong and Maria del Rio-Chanona and Dániel Kondor and Pieter Francois and Ed Chalstrey and Jakob Zsambok and Dan Hoyer and Jenny Reddish and Jakob Hauser and Francisco-Javier Rodrigo-Ginés and Suchandra Datta and Maxwell Shepherd and Thom Kamphuis and Qizheng Zhang and Hyunjun Kim and Ruiji Sun and Jianzhu Yao and Franck Dernoncourt and Satyapriya Krishna and Sina Rismanchian and Bonan Pu and Francesco Pinto and Yingheng Wang and Kumar Shridhar and Kalon J. Overholt and Glib Briia and Hieu Nguyen and David and Soler Bartomeu and Tony CY Pang and Adam Wecker and Yifan Xiong and Fanfei Li and Lukas S. Huber and Joshua Jaeger and Romano De Maddalena and Xing Han Lù and Yuhui Zhang and Claas Beger and Patrick Tser Jern Kon and Sean Li and Vivek Sanker and Ming Yin and Yihao Liang and Xinlu Zhang and Ankit Agrawal and Li S. Yifei and Zechen Zhang and Mu Cai and Yasin Sonmez and Costin Cozianu and Changhao Li and Alex Slen and Shoubin Yu and Hyun Kyu Park and Gabriele Sarti and Marcin Briański and Alessandro Stolfo and Truong An Nguyen and Mike Zhang and Yotam Perlitz and Jose Hernandez-Orallo and Runjia Li and Amin Shabani and Felix Juefei-Xu and Shikhar Dhingra and Orr Zohar and My Chiffon Nguyen and Alexander Pondaven and Abdurrahim Yilmaz and Xuandong Zhao and Chuanyang Jin and Muyan Jiang and Stefan Todoran and Xinyao Han and Jules Kreuer and Brian Rabern and Anna Plassart and Martino Maggetti and Luther Yap and Robert Geirhos and Jonathon Kean and Dingsu Wang and Sina Mollaei and Chenkai Sun and Yifan Yin and Shiqi Wang and Rui Li and Yaowen Chang and Anjiang Wei and Alice Bizeul and Xiaohan Wang and Alexandre Oliveira Arrais and Kushin Mukherjee and Jorge Chamorro-Padial and Jiachen Liu and Xingyu Qu and Junyi Guan and Adam Bouyamourn and Shuyu Wu and Martyna Plomecka and Junda Chen and Mengze Tang and Jiaqi Deng and Shreyas Subramanian and Haocheng Xi and Haoxuan Chen and Weizhi Zhang and Yinuo Ren and Haoqin Tu and Sejong Kim and Yushun Chen and Sara Vera Marjanović and Junwoo Ha and Grzegorz Luczyna and Jeff J. Ma and Zewen Shen and Dawn Song and Cedegao E. Zhang and Zhun Wang and Gaël Gendron and Yunze Xiao and Leo Smucker and Erica Weng and Kwok Hao Lee and Zhe Ye and Stefano Ermon and Ignacio D. Lopez-Miguel and Theo Knights and Anthony Gitter and Namkyu Park and Boyi Wei and Hongzheng Chen and Kunal Pai and Ahmed Elkhanany and Han Lin and Philipp D. Siedler and Jichao Fang and Ritwik Mishra and Károly Zsolnai-Fehér and Xilin Jiang and Shadab Khan and Jun Yuan and Rishab Kumar Jain and Xi Lin and Mike Peterson and Zhe Wang and Aditya Malusare and Maosen Tang and Isha Gupta and Ivan Fosin and Timothy Kang and Barbara Dworakowska and Kazuki Matsumoto and Guangyao Zheng and Gerben Sewuster and Jorge Pretel Villanueva and Ivan Rannev and Igor Chernyavsky and Jiale Chen and Deepayan Banik and Ben Racz and Wenchao Dong and Jianxin Wang and Laila Bashmal and Duarte V. Gonçalves and Wei Hu and Kaushik Bar and Ondrej Bohdal and Atharv Singh Patlan and Shehzaad Dhuliawala and Caroline Geirhos and Julien Wist and Yuval Kansal and Bingsen Chen and Kutay Tire and Atak Talay Yücel and Brandon Christof and Veerupaksh Singla and Zijian Song and Sanxing Chen and Jiaxin Ge and Kaustubh Ponkshe and Isaac Park and Tianneng Shi and Martin Q. Ma and Joshua Mak and Sherwin Lai and Antoine Moulin and Zhuo Cheng and Zhanda Zhu and Ziyi Zhang and Vaidehi Patil and Ketan Jha and Qiutong Men and Jiaxuan Wu and Tianchi Zhang and Bruno Hebling Vieira and Alham Fikri Aji and Jae-Won Chung and Mohammed Mahfoud and Ha Thi Hoang and Marc Sperzel and Wei Hao and Kristof Meding and Sihan Xu and Vassilis Kostakos and Davide Manini and Yueying Liu and Christopher Toukmaji and Jay Paek and Eunmi Yu and Arif Engin Demircali and Zhiyi Sun and Ivan Dewerpe and Hongsen Qin and Roman Pflugfelder and James Bailey and Johnathan Morris and Ville Heilala and Sybille Rosset and Zishun Yu and Peter E. Chen and Woongyeong Yeo and Eeshaan Jain and Ryan Yang and Sreekar Chigurupati and Julia Chernyavsky and Sai Prajwal Reddy and Subhashini Venugopalan and Hunar Batra and Core Francisco Park and Hieu Tran and Guilherme Maximiano and Genghan Zhang and Yizhuo Liang and Hu Shiyu and Rongwu Xu and Rui Pan and Siddharth Suresh and Ziqi Liu and Samaksh Gulati and Songyang Zhang and Peter Turchin and Christopher W. Bartlett and Christopher R. Scotese and Phuong M. Cao and Ben Wu and Jacek Karwowski and Davide Scaramuzza and Aakaash Nattanmai and Gordon McKellips and Anish Cheraku and Asim Suhail and Ethan Luo and Marvin Deng and Jason Luo and Ashley Zhang and Kavin Jindel and Jay Paek and Kasper Halevy and Allen Baranov and Michael Liu and Advaith Avadhanam and David Zhang and Vincent Cheng and Brad Ma and Evan Fu and Liam Do and Joshua Lass and Hubert Yang and Surya Sunkari and Vishruth Bharath and Violet Ai and James Leung and Rishit Agrawal and Alan Zhou and Kevin Chen and Tejas Kalpathi and Ziqi Xu and Gavin Wang and Tyler Xiao and Erik Maung and Sam Lee and Ryan Yang and Roy Yue and Ben Zhao and Julia Yoon and Sunny Sun and Aryan Singh and Ethan Luo and Clark Peng and Tyler Osbey and Taozhi Wang and Daryl Echeazu and Hubert Yang and Timothy Wu and Spandan Patel and Vidhi Kulkarni and Vijaykaarti Sundarapandiyan and Ashley Zhang and Andrew Le and Zafir Nasim and Srikar Yalam and Ritesh Kasamsetty and Soham Samal and Hubert Yang and David Sun and Nihar Shah and Abhijeet Saha and Alex Zhang and Leon Nguyen and Laasya Nagumalli and Kaixin Wang and Alan Zhou and Aidan Wu and Jason Luo and Anwith Telluri and Summer Yue and Alexandr Wang and Dan Hendrycks},
      year={2025},
      eprint={2501.14249},
      archivePrefix={arXiv},
      primaryClass={cs.LG},
      url={https://arxiv.org/abs/2501.14249}, 
}

\section{Appendix}

\subsection{Syllogism and Categorical Propositions}

The core structure of the syllogism was first systematically articulated by the ancient Greek philosopher Aristotle (384--322~BCE) in the \emph{Organon}. He defines a syllogism as a form of reasoning in which the conclusion follows necessarily from the premises, and it is standardly analyzed as involving a major premise, a minor premise, and a conclusion. A standard-form categorical syllogism is built from three core components \cite{Aristotle, Copi2014Logic}:
\begin{itemize}
    \item \textbf{Three Terms:}
    \begin{itemize}
        \item The \textbf{major term (P)} is the predicate of the conclusion.
        \item The \textbf{minor term (S)} is the subject of the conclusion.
        \item The \textbf{middle term (M)} appears in both premises but not in the conclusion.
    \end{itemize}
    \item \textbf{Three Propositions:}
    \begin{itemize}
        \item The \textbf{major premise} contains the major term (P) and the middle term (M).
        \item The \textbf{minor premise} contains the minor term (S) and the middle term (M).
        \item The \textbf{conclusion} links the minor term (S) to the major term (P).
    \end{itemize}
\end{itemize}

In Aristotelian syllogistic logic (traditional logic), categorical propositions are divided into  \textbf{four standard forms}:
\begin{itemize}
  \item \textbf{A}-proposition (universal affirmative), of the form ``All $S$ are $P$,''
        e.g., ``All humans are mortal''.
  \item \textbf{E}-proposition (universal negative), of the form ``No $S$ are $P$,''
        e.g., ``No humans are perfect''.
  \item \textbf{I}-proposition (particular affirmative), of the form ``Some $S$ are $P$,''
        e.g., ``Some humans are healthy''.
  \item \textbf{O}-proposition (particular negative), of the form ``Some $S$ are not $P$,''
        e.g., ``Some humans are not healthy''.
\end{itemize}

In a categorical syllogism, both premises and the conclusion are propositions of these four types \cite{Suppes1957-SUPITL}. In the traditional (Aristotelian) interpretation, the truth of a universal proposition is taken to imply the truth of its corresponding particular proposition \cite{rasooli-tetrault-2015}. This assumption licenses, for example, subalternation from an A-proposition to the corresponding I-proposition; e.g., from ``All humans are mortal'' one may infer ``Some humans are mortal.''

In contrast, George Boole, a nineteenth-century English mathematician, argued that we cannot infer the truth of a particular proposition from the truth of its corresponding universal proposition, because every particular proposition asserts the existence of its subject class. If a universal proposition permitted us to infer the corresponding particular, then "All leprechauns wear little green hats" would license the inference that some leprechauns do, which would imply that there really are leprechauns \cite{Boole1854}. Thus, under modern logic (the Boolean interpretation), a universal proposition (an A- or E-proposition) is understood as stating only, for example, "If there is such a thing as a leprechaun, it wears a little green hat," not that any leprechauns actually exist.

\subsection{Formalization of Categorical Propositions}
\label{24valid}

In the main text, we adopt the formalization of modern logic (Boolean), which reinterprets categorical propositions as quantified formulas. Throughout, by modern logic we mean the Boolean interpretation (no existential import for universals), expressed using standard quantified notation. The typical correspondences are as follows:

\begin{table}[H]
\centering
\small
\setlength{\tabcolsep}{0.5pt}
\renewcommand{\arraystretch}{1.0}

\newcommand{\Strut}{\rule{0pt}{2.8ex}\rule[-1.2ex]{0pt}{0pt}}

\newlength{\cwa}\setlength{\cwa}{0.30\columnwidth}
\newlength{\cwb}\setlength{\cwb}{0.30\columnwidth}
\newlength{\cwc}\setlength{\cwc}{0.38\columnwidth}

\newcolumntype{L}[1]{>{\raggedright\arraybackslash}m{#1}}

\begin{tabular}{|L{\cwa}|L{\cwb}|L{\cwc}|}
\hline
\Strut\makecell[l]{\textbf{Categorical}\\\textbf{Proposition}} &
\Strut\makecell[l]{\textbf{Formalization in}\\\textbf{modern logic}} &
\Strut\makecell[l]{\textbf{Explanation}} \\
\hline
\Strut All $S$ are $P$ &
\Strut $\forall x\, (Sx \rightarrow Px)$ &
\Strut For all $x$, if $x$ is $S$, then $x$ is $P$. \\
\hline
\Strut No $S$ are $P$ &
\Strut \makecell[l]{$\forall x\, (Sx \rightarrow \neg Px)$\\or\\$\neg \exists x\, (Sx \wedge Px)$} &
\Strut For all $x$, if $x$ is $S$, then $x$ is not $P$; equivalently, there does not exist any $x$ such that $x$ is both $S$ and $P$. \\
\hline
\Strut Some $S$ are $P$ &
\Strut $\exists x\, (Sx \wedge Px)$ &
\Strut There exists at least one $x$ such that $x$ is $S$ and $x$ is $P$. \\
\hline
\Strut Some $S$ are not $P$ &
\Strut $\exists x\, (Sx \wedge \neg Px)$ &
\Strut There exists at least one $x$ such that $x$ is $S$ and $x$ is not $P$. \\
\hline
\end{tabular}

\label{tab:categorical-propositions}
\end{table}

Under the standard semantics of modern logic~\cite{enderton-math-logic}, if the extension of $S$ is empty, then
$\exists x(Sx \land Px)$ is false, whereas $\forall x(Sx \rightarrow Px)$ is
vacuously true. Consequently, from $\forall x(Sx \rightarrow Px)$ and
$\forall x(Px \rightarrow Qx)$ one cannot derive $\exists x(Sx \land Qx)$ unless
one adds an extra existence assumption (e.g., $\exists x\, Sx$). For this reason, as shown in Table~\ref{tab:syl-15-9}, among the 24 standard syllogistic forms treated as valid in traditional logic, 9 are not valid in general under the semantics of modern logic, because their correctness depends on existential import  \cite{Copi2014Logic}.

\subsection{A Modern-Logic Derivation of Barbara}

Take the syllogism Barbara (mood AAA in the first figure) as an example:
{\small
\begin{align*}
\text{Major premise:} &\quad \text{All } M \text{ are } P \;\to\; \forall x (Mx \rightarrow Px),\\
\text{Minor premise:} &\quad \text{All } S \text{ are } M \;\to\; \forall x (Sx \rightarrow Mx),\\
\text{Conclusion:} &\quad \text{All } S \text{ are } P \;\to\; \forall x (Sx \rightarrow Px).
\end{align*}
}
Under modern logic, the validity of this inference can be demonstrated by a formal derivation (e.g., in natural deduction):
{\small
\begin{enumerate}
  \item $\forall x (Mx \rightarrow Px)$ \hfill [Major premise]
  \item $\forall x (Sx \rightarrow Mx)$ \hfill [Minor premise]
  \item Assume an arbitrary $a$. \hfill [Arbitrary individual]
  \item $Sa \rightarrow Ma$ \hfill [from 2, $\forall$-elim]
  \item $Ma \rightarrow Pa$ \hfill [from 1, $\forall$-elim]
  \item $Sa \rightarrow Pa$ \hfill [from 4, 5]
  \item $\forall x (Sx \rightarrow Px)$ \hfill [from 3--6, $\forall$-intro]
\end{enumerate}
}

This example shows that traditional syllogistic reasoning can be formalized and
verified within modern logic.

\begin{table}[t]
\centering
\begin{tabular}{llll}
\hline
\textbf{Name} & \textbf{Mood} & \textbf{Figure} & \textbf{Validity} \\
\hline
Barbara   & AAA & I   & Valid in Both \\
Celarent  & EAE & I   & Valid in Both \\
Darii     & AII & I   & Valid in Both \\
Ferio     & EIO & I   & Valid in Both \\
Barbari   & AAI & I   & Traditional only  \\
Celaront  & EAO & I   & Traditional only  \\
\hline
Cesare    & EAE & II  & Valid in Both \\
Camestres & AEE & II  & Valid in Both \\
Festino   & EIO & II  & Valid in Both \\
Baroco    & AOO & II  & Valid in Both \\
Cesaro    & EAO & II  & Traditional only  \\
Camestrop & AEO & II  & Traditional only  \\ 
\hline
Darapti   & AAI & III & Traditional only  \\
Disamis   & IAI & III & Valid in Both \\
Datisi    & AII & III & Valid in Both \\
Felapton  & EAO & III & Traditional only  \\
Bocardo   & OAO & III & Valid in Both \\
Ferison   & EIO & III & Valid in Both \\
\hline
Bamalip   & AAI & IV  & Traditional only  \\
Camenes   & AEE & IV  & Valid in Both \\
Dimaris   & IAI & IV  & Valid in Both \\ 
Calemop   & AEO & IV  & Traditional only  \\ 
Fesapo    & EAO & IV  & Traditional only  \\
Fresison  & EIO & IV  & Valid in Both \\
\hline
\end{tabular}
\caption{The 15+9 Distinction of Valid Syllogistic Forms (Traditional logic vs.\ Modern logic)}
\label{tab:syl-15-9}
\end{table}

\subsection{Data Construction}
\label{data}
The dataset construction process follows a rigorously structured three-stage pipeline: (1) Diverse Topic Seeding, (2) Closed-Loop Generation and Verification, and (3) Triplet Completion and Relational Validation. Each stage is designed to build upon the previous one, progressively refining the quality and logical richness of the resulting data.

\paragraph{Diverse Topic Seeding.}
To ensure broad topical coverage and prevent semantic bias toward common or overrepresented domains, the process begins with a topic seeding stage. A predefined set of meta-domains spanning natural sciences, engineering, social sciences, and the humanities is used as the high-level taxonomy. For each meta-domain, a Topic Generation Agent is prompted to produce a set of concrete and verifiable subfields or research directions that exist in reality. The outcome is a diverse and fine-grained collection of topics, each serving as a contextual anchor for subsequent concept generation. This stage establishes semantic breadth and ensures that reasoning patterns later derived from the dataset are not constrained to narrow disciplinary vocabularies.

\paragraph{Closed-Loop Generation and Verification}
At the core of the dataset construction process lies the closed-loop generation and verification stage, which establishes the factual and semantic foundation of both non-empty and empty sets for the minor term $(S)$ within each syllogistic structure. This stage guarantees that generated concepts are not only syntactically well-formed but also ontologically consistent with their designated existential category. Two complementary generation objectives are defined: \textit{non-empty} concepts, which correspond to empirically verifiable entities in the real world, and \textit{empty} concepts, which remain logically coherent while representing categories with no real-world instantiation.

For each topic obtained from the previous stage, an iterative "generate, verify, feedback" loop is executed. The Generator agent first produces a candidate concept $(S)$ that satisfies the existential target of the current data subset. The candidate is then evaluated by a panel of independent Validator agents, each performing an autonomous factuality assessment and issuing a categorical verdict("non-empty" and "empty") accompanied by explanatory reasoning and indicative verification paths. A concept advances only when all validators unanimously agree on the verdict corresponding to the intended generation type, confirming either its empirical existence (for non-empty cases) or its verified non-existence (for empty cases). If disagreement arises, the system consolidates validator feedback into a unified critique, which is returned to the Generator in the next iteration to guide conceptual refinement. Through this iterative feedback-driven process, the framework produces two balanced sets of high-confidence concepts that jointly represent existentially positive and negative categories of reality.

\paragraph{Triplet Completion and Relational Validation}
After obtaining a validated non-empty or empty concept $(S)$, the final stage completes the triplet structure by generating the corresponding middle $(M)$ and major $(P)$ terms. The Triplet Generator agent constructs the set $(S, M, P)$ under strict constraints ensuring that all three concepts belong to a coherent semantic frame amenable to syllogistic reasoning. The agent is explicitly instructed to avoid trivial or hierarchical relations such as synonymy or direct subclass relationships (e.g., \textit{poodle} $\rightarrow$ \textit{dog} $\rightarrow$ \textit{animal}), instead favoring more nuanced logical relations grounded in attribute overlap, contextual differentiation, or mechanistic contrast. To enforce this non-triviality constraint, each triplet undergoes an additional Relational Validation phase. Here, Validator agents examine whether deterministic subsumption or equivalence relations exist among the three concepts. A triplet is finalized only if it passes this logical consistency test, confirming its suitability for constructing non-trivial reasoning scenarios.

\paragraph{Syllogistic Data Realization}
Upon successful generation and validation of all $(S, M, P)$ triplets, the final step transforms these verified conceptual structures into complete syllogistic reasoning instances. Each triplet serves as a semantic scaffold that is systematically mapped onto the twenty-four canonical syllogistic mood–figure templates formalized in Aristotelian logic. By substituting the generated concepts into these templates, the system produces a diverse collection of categorical syllogisms encompassing universal affirmatives, particulars, and negatives across multiple structural figures. This synthesis ensures that every syllogistic instance conforms to formal logical syntax while remaining grounded in verifiable semantic content. The resulting corpus thus unifies traditional deductive structures with empirically meaningful concepts, providing a rigorous benchmark for evaluating machine reasoning under both semantic authenticity and logical validity.

\subsection{The precision and recall metrics}
\label{precision_recall}

Under modern logic, 15 syllogisms are regarded as valid, while the remaining 9 are invalid. We treat valid forms as positive samples $(P)$ and invalid forms as negative samples $(N)$.
We define the precision and recall metrics of valid and invalid syllogisms $\mathrm{pre}_V, \mathrm{rec}_V, \mathrm{pre}_I, \mathrm{rec}_I$ as follow:

\[
\mathrm{pre}_V = \frac{TP}{TP+FP}
\]
\[
\mathrm{rec}_V = \frac{TP}{TP+FN}
\]
\[
\mathrm{pre}_I = \frac{TN}{TN+FN}
\]
\[
\mathrm{rec}_I =\frac{TN}{TN+FP}
\]

\subsection{The results of experiments}

\subsubsection{The detailed results of closed-source models}
\label{app_closed_source}
The detailed results of closed-source models are shown in Tab.\ref{tab:closed_source_by_lang_concept_detailed}. GPT-5 and GPT-o3 exhibit the extreme modern logic tendency.

\begin{table*}[t]
\centering
\resizebox{\textwidth}{!}{%
\begin{tabular}{l|ccccccc|ccccccc|ccccccc|ccccccc}
\toprule
Model & \multicolumn{7}{c}{ZH+} & \multicolumn{7}{c}{ZH-} & \multicolumn{7}{c}{EN+} & \multicolumn{7}{c}{EN-} \\
\cmidrule(lr){2-8}
\cmidrule(lr){9-15}
\cmidrule(lr){16-22}
\cmidrule(lr){23-29}
  & $\text{Acc}_{t}$ & $\text{Acc}_{m}$ & Cons & $\mathrm{pre}_V$ & $\mathrm{rec}_V$ & $\mathrm{pre}_I$ & $\mathrm{rec}_I$ & $\text{Acc}_{t}$ & $\text{Acc}_{m}$ & Cons & $\mathrm{pre}_V$ & $\mathrm{rec}_V$ & $\mathrm{pre}_I$ & $\mathrm{rec}_I$ & $\text{Acc}_{t}$ & $\text{Acc}_{m}$ & Cons & $\mathrm{pre}_V$ & $\mathrm{rec}_V$ & $\mathrm{pre}_I$ & $\mathrm{rec}_I$ & $\text{Acc}_{t}$ & $\text{Acc}_{m}$ & Cons & $\mathrm{pre}_V$ & $\mathrm{rec}_V$ & $\mathrm{pre}_I$ & $\mathrm{rec}_I$ \\
\midrule
Claude-3.7-Sonnet & 85.29 & 76.54 & 45.83 & 72.89 & 99.47 & 97.73 & 38.33 & 90.46 & 71.71 & 50.00 & 68.91 & 99.73 & 98.25 & 25.00 & 70.33 & 92.00 & 54.17 & 88.74 & 99.87 & 99.72 & 78.89 & 73.08 & 89.42 & 62.50 & 85.52 & 100.00 & 100.00 & 71.78 \\
Claude-4.5-Sonnet & 81.38 & 81.12 & 62.50 & 76.80 & 100.00 & 100.00 & 49.67 & 93.96 & 68.57 & 62.50 & 66.55 & 100.00 & 100.00 & 16.13 & 70.01 & 92.52 & 66.67 & 89.32 & 100.00 & 100.00 & 80.04 & 84.11 & 78.40 & 62.50 & 74.32 & 100.00 & 100.00 & 42.38 \\
Gemini-2.5-Pro & 71.92 & 89.33 & 29.17 & 86.04 & 99.07 & 97.92 & 73.22 & 76.17 & 83.50 & 25.00 & 80.20 & 97.73 & 94.06 & 59.78 & 65.17 & 97.33 & 70.83 & 95.91 & 100.00 & 100.00 & 92.89 & 72.92 & 89.50 & 58.33 & 85.66 & 100.00 & 100.00 & 72.11 \\
Gemini-3-Pro-Preview & 73.11 & 89.20 & 54.17 & 85.35 & 99.87 & 99.69 & 71.44 & 99.00 & 63.48 & 66.67 & 63.12 & 100.00 & 100.00 & 2.67 & 63.48 & 99.00 & 79.17 & 98.42 & 100.00 & 100.00 & 97.33 & 98.41 & 64.02 & 70.83 & 63.44 & 100.00 & 100.00 & 4.22 \\
GPT-4o-2024-11-20 & 93.17 & 68.42 & 41.67 & 66.68 & 99.53 & 95.57 & 16.85 & 96.17 & 65.71 & 50.00 & 64.73 & 99.73 & 95.40 & 9.25 & 93.33 & 68.75 & 50.00 & 66.83 & 99.87 & 98.71 & 17.08 & 94.04 & 67.83 & 50.00 & 66.15 & 99.60 & 95.74 & 15.02 \\
GPT-4.1-2025-04-14 & 80.38 & 80.04 & 33.33 & 76.46 & 98.33 & 94.69 & 49.56 & 85.08 & 76.67 & 45.83 & 73.02 & 99.40 & 97.49 & 38.78 & 80.04 & 82.38 & 58.33 & 78.03 & 99.93 & 99.79 & 53.11 & 81.54 & 80.96 & 62.50 & 76.65 & 100.00 & 100.00 & 49.22 \\
GPT-o3 & 62.38 & 99.54 & 87.50 & 99.73 & 99.60 & 99.33 & 99.56 & 62.58 & 99.92 & 91.67 & 99.87 & 100.00 & 100.00 & 99.78 & 62.50 & 100.00 & 100.00 & 100.00 & 100.00 & 100.00 & 100.00 & 62.58 & 99.92 & 95.83 & 99.87 & 100.00 & 100.00 & 99.78 \\

GPT-5-2025-08-07 & 62.53 & 100.00 & 100.00 & 100.00 & 100.00 & 100.00 & 100.00 & 62.33 & 100.00 & 100.00 & 100.00 & 100.00 & 100.00 & 100.00 & 62.53 & 100.00 & 100.00 & 100.00 & 100.00 & 100.00 & 100.00 & 62.40 & 100.00 & 100.00 & 100.00 & 100.00 & 100.00 & 100.00 \\

\bottomrule
\end{tabular}
}
\caption{The detailed results of closed-source models.}
\label{tab:closed_source_by_lang_concept_detailed}
\end{table*}

\subsubsection{The detailed results of experiment with Prior-check prompt}
\label{sec:appendix_main_results}

The results are shown in Tab.\ref{tab:open_source_by_lang_concept_detailed}.

\begin{table*}[t]
\centering
\resizebox{\textwidth}{!}{%
\begin{tabular}{l|ccccccc|ccccccc|ccccccc|ccccccc}
\toprule
Model & \multicolumn{7}{c}{ZH+} & \multicolumn{7}{c}{ZH-} & \multicolumn{7}{c}{EN+} & \multicolumn{7}{c}{EN-} \\
\cmidrule(lr){2-8}
\cmidrule(lr){9-15}
\cmidrule(lr){16-22}
\cmidrule(lr){23-29}
  & $\text{Acc}_{t}$ & $\text{Acc}_{m}$ & Cons & $\mathrm{pre}_V$ & $\mathrm{rec}_V$ & $\mathrm{pre}_I$ & $\mathrm{rec}_I$
  & $\text{Acc}_{t}$ & $\text{Acc}_{m}$ & Cons & $\mathrm{pre}_V$ & $\mathrm{rec}_V$ & $\mathrm{pre}_I$ & $\mathrm{rec}_I$
  & $\text{Acc}_{t}$ & $\text{Acc}_{m}$ & Cons & $\mathrm{pre}_V$ & $\mathrm{rec}_V$ & $\mathrm{pre}_I$ & $\mathrm{rec}_I$
  & $\text{Acc}_{t}$ & $\text{Acc}_{m}$ & Cons & $\mathrm{pre}_V$ & $\mathrm{rec}_V$ & $\mathrm{pre}_I$ & $\mathrm{rec}_I$ \\
\midrule
Qwen3-0.6B & 100.00 & 62.50 & 100.00 & 62.50 & 100.00 & 0.00 & 0.00 & 99.96 & 62.46 & 95.83 & 62.48 & 99.93 & 0.00 & 0.00 & 100.00 & 62.50 & 100.00 & 62.50 & 100.00 & 0.00 & 0.00 & 100.00 & 62.50 & 100.00 & 62.50 & 100.00 & 0.00 & 0.00 \\
Qwen3-0.6B-Thinking & 94.71 & 61.04 & 4.17 & 62.43 & 94.60 & 36.22 & 5.11 & 92.96 & 61.12 & 16.67 & 62.71 & 93.27 & 40.24 & 7.56 & 86.67 & 60.25 & 0.00 & 63.12 & 87.53 & 41.56 & 14.78 & 88.33 & 61.75 & 4.17 & 63.73 & 90.07 & 46.79 & 14.56 \\
Qwen3-1.7B & 97.00 & 62.42 & 50.00 & 62.84 & 97.53 & 48.61 & 3.89 & 95.58 & 60.92 & 37.50 & 62.25 & 95.20 & 32.08 & 3.78 & 75.21 & 59.71 & 16.67 & 64.76 & 77.93 & 44.37 & 29.33 & 35.17 & 47.58 & 4.17 & 64.34 & 36.20 & 38.50 & 66.56 \\
Qwen3-1.7B-Thinking & 92.92 & 67.67 & 29.17 & 66.23 & 98.47 & 86.47 & 16.33 & 94.29 & 67.71 & 50.00 & 66.02 & 99.60 & 95.62 & 14.56 & 91.62 & 70.54 & 54.17 & 68.03 & 99.73 & 98.01 & 21.89 & 91.96 & 70.29 & 58.33 & 67.83 & 99.80 & 98.45 & 21.11 \\

Qwen3-4B & 92.46 & 67.12 & 45.83 & 66.02 & 97.67 & 80.66 & 16.22 & 94.46 & 67.04 & 54.17 & 65.64 & 99.20 & 90.98 & 13.44 & 85.79 & 61.62 & 4.17 & 64.06 & 87.93 & 46.92 & 17.78 & 93.50 & 61.67 & 12.50 & 62.92 & 94.13 & 43.59 & 7.56 \\
Qwen3-4B-Thinking & 82.54 & 79.96 & 62.50 & 75.72 & 100.00 & 100.00 & 46.56 & 85.33 & 77.08 & 58.33 & 73.19 & 99.93 & 99.72 & 39.00 & 83.62 & 78.88 & 66.67 & 74.74 & 100.00 & 100.00 & 43.67 & 84.92 & 77.58 & 62.50 & 73.60 & 100.00 & 100.00 & 40.22 \\
Qwen3-8B & 94.12 & 67.46 & 33.33 & 65.91 & 99.27 & 92.20 & 14.44 & 96.67 & 65.42 & 62.50 & 64.44 & 99.67 & 93.75 & 8.33 & 85.46 & 69.58 & 4.17 & 68.80 & 94.07 & 74.43 & 28.81 & 86.71 & 64.62 & 0.00 & 65.64 & 91.07 & 57.99 & 20.56 \\
Qwen3-8B-Thinking & 67.83 & 94.50 & 54.17 & 92.01 & 99.87 & 99.74 & 85.56 & 71.62 & 90.88 & 62.50 & 87.26 & 100.00 & 100.00 & 75.67 & 64.83 & 97.67 & 75.00 & 96.40 & 100.00 & 100.00 & 93.78 & 65.29 & 97.21 & 66.67 & 95.72 & 100.00 & 100.00 & 92.56 \\

Qwen3-14B & 97.75 & 64.50 & 66.67 & 63.81 & 99.80 & 94.44 & 5.67 & 99.25 & 63.25 & 87.50 & 62.97 & 100.00 & 100.00 & 2.00 & 87.12 & 70.96 & 25.00 & 69.20 & 96.47 & 82.85 & 28.44 & 91.58 & 68.08 & 20.83 & 66.70 & 97.73 & 83.17 & 18.67 \\
Qwen3-14B-Thinking & 72.96 & 89.54 & 62.50 & 85.67 & 100.00 & 100.00 & 72.11 & 76.50 & 86.00 & 66.67 & 81.70 & 100.00 & 100.00 & 62.67 & 74.92 & 87.50 & 58.33 & 83.37 & 99.93 & 99.83 & 66.78 & 77.92 & 84.50 & 58.33 & 80.16 & 99.93 & 99.81 & 58.78 \\
Qwen3-32B & 91.67 & 70.33 & 58.33 & 67.91 & 99.60 & 97.00 & 21.56 & 95.54 & 66.96 & 75.00 & 65.42 & 100.00 & 100.00 & 11.89 & 91.00 & 70.50 & 45.83 & 68.13 & 99.20 & 94.44 & 22.67 & 93.88 & 68.46 & 54.17 & 66.49 & 99.87 & 98.64 & 16.11 \\
Qwen3-32B-Thinking & 82.21 & 80.29 & 62.50 & 76.03 & 100.00 & 100.00 & 47.44 & 85.75 & 76.75 & 62.50 & 72.89 & 100.00 & 100.00 & 38.00 & 77.96 & 84.50 & 62.50 & 80.17 & 100.00 & 100.00 & 58.73 & 80.38 & 82.08 & 62.50 & 77.76 & 100.00 & 100.00 & 52.28 \\

Qwen3-30B-A3B-Instruct & 66.58 & 95.83 & 70.83 & 93.80 & 99.93 & 99.88 & 89.00 & 71.96 & 90.54 & 66.67 & 86.86 & 100.00 & 100.00 & 74.78 & 64.00 & 98.50 & 75.00 & 97.66 & 100.00 & 100.00 & 96.00 & 66.71 & 95.71 & 66.67 & 93.63 & 99.93 & 99.87 & 88.67 \\
Qwen3-30B-A3B-Thinking & 69.17 & 93.33 & 62.50 & 90.36 & 100.00 & 100.00 & 82.22 & 71.50 & 91.00 & 62.50 & 87.41 & 100.00 & 100.00 & 76.00 & 67.71 & 86.12 & 16.67 & 85.91 & 93.07 & 86.58 & 74.56 & 70.00 & 84.08 & 8.33 & 83.27 & 93.27 & 85.97 & 68.78 \\

Qwen3-next-80B-A3B-Instruct & 65.58 & 96.92 & 66.67 & 95.30 & 100.00 & 100.00 & 91.78 & 70.08 & 92.42 & 66.67 & 89.18 & 100.00 & 100.00 & 79.78 & 62.71 & 99.62 & 70.83 & 99.53 & 99.87 & 99.78 & 99.22 & 64.38 & 98.12 & 62.50 & 97.09 & 100.00 & 100.00 & 95.00 \\
Qwen3-next-80B-A3B-Thinking & 62.71 & 99.79 & 83.33 & 99.67 & 100.00 & 100.00 & 99.44 & 63.08 & 99.42 & 79.17 & 99.08 & 100.00 & 100.00 & 98.44 & 62.88 & 98.96 & 50.00 & 98.87 & 99.47 & 99.10 & 98.11 & 62.96 & 99.38 & 75.00 & 99.14 & 99.87 & 99.78 & 98.56 \\
Qwen3-235B-A22B-Instruct & 66.17 & 96.33 & 66.67 & 94.46 & 100.00 & 100.00 & 90.22 & 67.83 & 94.67 & 66.67 & 92.14 & 100.00 & 100.00 & 85.78 & 62.54 & 99.88 & 87.50 & 99.87 & 99.93 & 99.89 & 99.78 & 62.71 & 99.79 & 83.33 & 99.67 & 100.00 & 100.00 & 99.44 \\
Qwen3-235B-A22B-Thinking & 62.71 & 99.79 & 83.33 & 99.67 & 100.00 & 100.00 & 99.44 & 62.88 & 99.62 & 83.33 & 99.40 & 100.00 & 100.00 & 99.00 & 64.75 & 97.75 & 62.50 & 96.53 & 100.00 & 100.00 & 94.00 & 63.08 & 99.42 & 70.83 & 99.08 & 100.00 & 100.00 & 98.44 \\

Gemma-3-1B-IT & 87.96 & 53.29 & 0.00 & 58.98 & 83.00 & 11.76 & 3.78 & 77.62 & 51.71 & 0.00 & 59.15 & 73.47 & 25.88 & 15.44 & 90.29 & 57.54 & 0.00 & 61.10 & 88.27 & 24.46 & 6.33 & 86.71 & 57.54 & 0.00 & 61.56 & 85.40 & 31.35 & 11.11 \\
Gemma-3-4B-IT & 94.46 & 63.38 & 16.67 & 63.70 & 96.27 & 57.89 & 8.56 & 77.88 & 63.54 & 0.00 & 66.72 & 83.13 & 52.35 & 30.89 & 95.00 & 63.08 & 12.50 & 63.46 & 96.47 & 55.83 & 7.44 & 94.79 & 64.38 & 25.00 & 64.18 & 97.33 & 68.00 & 9.44 \\
Gemma-3-12B-IT & 98.54 & 63.38 & 41.67 & 63.13 & 99.53 & 80.00 & 3.11 & 98.96 & 62.88 & 45.83 & 62.82 & 99.47 & 68.00 & 1.89 & 93.67 & 63.42 & 20.83 & 63.83 & 95.67 & 57.24 & 9.67 & 92.38 & 64.96 & 20.83 & 64.86 & 95.87 & 66.12 & 13.44 \\
Gemma-3-27B-IT & 95.33 & 62.00 & 16.67 & 62.85 & 95.87 & 44.64 & 5.56 & 94.17 & 61.58 & 20.83 & 62.79 & 94.60 & 42.14 & 6.56 & 96.54 & 65.71 & 50.00 & 64.61 & 99.80 & 96.39 & 8.89 & 95.96 & 66.54 & 66.67 & 65.13 & 100.00 & 100.00 & 10.78 \\
Llama3-8B-Instruct & 75.12 & 60.21 & 0.00 & 65.17 & 78.49 & 45.61 & 30.07 & 63.29 & 53.79 & 0.00 & 62.87 & 63.67 & 38.14 & 37.33 & 50.25 & 56.88 & 0.00 & 69.32 & 55.73 & 44.34 & 58.84 & 47.42 & 51.83 & 0.00 & 65.11 & 49.43 & 39.89 & 55.89 \\
Llama3-70B-Instruct & 98.58 & 63.17 & 58.33 & 63.02 & 99.40 & 73.53 & 2.78 & 96.88 & 62.71 & 45.83 & 63.01 & 97.67 & 53.33 & 4.44 & 98.88 & 62.54 & 62.50 & 62.66 & 99.13 & 51.85 & 1.56 & 90.67 & 60.29 & 20.83 & 62.59 & 90.80 & 38.12 & 9.45 \\
Llama3.3-70B-Instruct & 96.08 & 65.92 & 58.33 & 64.79 & 99.60 & 93.62 & 9.78 & 97.88 & 63.96 & 62.50 & 63.52 & 99.47 & 84.31 & 4.78 & 99.08 & 63.00 & 87.50 & 62.87 & 99.67 & 77.27 & 1.89 & 99.12 & 63.38 & 79.17 & 63.05 & 100.00 & 100.00 & 2.33 \\
\bottomrule
\end{tabular}
}
\caption{The detailed results of open-source models.}
\label{tab:open_source_by_lang_concept_detailed}
\end{table*}

\subsubsection{The baseline experiment without the Prior-check prompt}
\label{sec:appendix_control_group}

The results are shown in Table~\ref{tab:by_lang_concept_cg_detailed}.

\begin{table*}[t]
\centering
\resizebox{\textwidth}{!}{%
\begin{tabular}{l|ccccccc|ccccccc|ccccccc|ccccccc}
\toprule
Model & \multicolumn{7}{c}{ZH+} & \multicolumn{7}{c}{ZH-} & \multicolumn{7}{c}{EN+} & \multicolumn{7}{c}{EN-} \\
\cmidrule(lr){2-8}
\cmidrule(lr){9-15}
\cmidrule(lr){16-22}
\cmidrule(lr){23-29}
  & $\text{Acc}_{t}$ & $\text{Acc}_{m}$ & Cons & $\mathrm{pre}_V$ & $\mathrm{rec}_V$ & $\mathrm{pre}_I$ & $\mathrm{rec}_I$ & $\text{Acc}_{t}$ & $\text{Acc}_{m}$ & Cons & $\mathrm{pre}_V$ & $\mathrm{rec}_V$ & $\mathrm{pre}_I$ & $\mathrm{rec}_I$ & $\text{Acc}_{t}$ & $\text{Acc}_{m}$ & Cons & $\mathrm{pre}_V$ & $\mathrm{rec}_V$ & $\mathrm{pre}_I$ & $\mathrm{rec}_I$ & $\text{Acc}_{t}$ & $\text{Acc}_{m}$ & Cons & $\mathrm{pre}_V$ & $\mathrm{rec}_V$ & $\mathrm{pre}_I$ & $\mathrm{rec}_I$ \\
\midrule
Qwen3-0.6B & 100.00 & 62.50 & 100.00 & 62.50 & 100.00 & 0.00 & 0.00 & 99.96 & 62.46 & 95.83 & 62.48 & 99.93 & 0.00 & 0.00 & 98.38 & 62.12 & 25.00 & 62.52 & 98.40 & 38.46 & 1.67 & 79.75 & 56.75 & 0.00 & 62.07 & 79.20 & 35.80 & 19.33 \\
Qwen3-0.6B-Thinking & 87.96 & 62.04 & 0.00 & 63.95 & 90.00 & 48.10 & 15.44 & 91.42 & 61.83 & 0.00 & 63.31 & 92.60 & 46.12 & 10.56 & 86.54 & 60.29 & 0.00 & 63.17 & 87.47 & 41.80 & 15.00 & 89.58 & 60.25 & 0.00 & 62.70 & 89.87 & 39.20 & 10.89 \\
Qwen3-1.7B & 93.04 & 63.04 & 20.83 & 63.73 & 94.87 & 53.89 & 10.00 & 94.50 & 62.75 & 29.17 & 63.36 & 95.80 & 52.27 & 7.67 & 93.00 & 62.67 & 37.50 & 63.53 & 94.53 & 51.19 & 9.56 & 70.96 & 52.46 & 0.00 & 60.54 & 68.73 & 32.71 & 25.33 \\
Qwen3-1.7B-Thinking & 93.46 & 67.46 & 37.50 & 66.03 & 98.73 & 87.90 & 15.33 & 94.00 & 67.75 & 54.17 & 66.09 & 99.40 & 93.75 & 15.00 & 93.38 & 67.88 & 29.17 & 66.27 & 99.00 & 90.57 & 16.00 & 93.83 & 68.58 & 62.50 & 66.56 & 99.93 & 99.32 & 16.33 \\
Qwen3-4B & 93.88 & 65.46 & 54.17 & 64.89 & 97.47 & 74.15 & 12.11 & 96.00 & 65.50 & 62.50 & 64.58 & 99.20 & 87.50 & 9.33 & 83.42 & 65.75 & 25.00 & 66.93 & 89.33 & 59.80 & 26.44 & 90.25 & 67.75 & 41.67 & 66.76 & 96.40 & 76.92 & 20.00 \\
Qwen3-4B-Thinking & 88.88 & 73.62 & 62.50 & 70.32 & 100.00 & 100.00 & 29.67 & 87.12 & 75.38 & 62.50 & 71.74 & 100.00 & 100.00 & 34.33 & 86.67 & 75.83 & 62.50 & 72.12 & 100.00 & 100.00 & 35.56 & 87.17 & 75.33 & 62.50 & 71.70 & 100.00 & 100.00 & 34.22 \\
Qwen3-8B & 92.42 & 68.00 & 41.67 & 66.50 & 98.33 & 86.26 & 17.44 & 95.54 & 65.46 & 45.83 & 64.63 & 98.80 & 83.18 & 9.89 & 88.38 & 66.25 & 29.17 & 66.29 & 93.73 & 66.19 & 20.47 & 91.54 & 66.29 & 33.33 & 65.73 & 96.27 & 72.41 & 16.33 \\
Qwen3-8B-Thinking & 75.83 & 86.67 & 62.50 & 82.42 & 100.00 & 100.00 & 64.44 & 79.17 & 83.33 & 62.50 & 78.95 & 100.00 & 100.00 & 55.56 & 75.42 & 87.08 & 62.50 & 82.87 & 100.00 & 100.00 & 65.56 & 73.96 & 88.54 & 62.50 & 84.51 & 100.00 & 100.00 & 69.44 \\

Qwen3-14B & 94.46 & 67.12 & 37.50 & 65.68 & 99.27 & 91.73 & 13.56 & 96.33 & 65.83 & 58.33 & 64.71 & 99.73 & 95.45 & 9.33 & 89.46 & 65.71 & 37.50 & 65.77 & 94.13 & 65.22 & 18.33 & 91.88 & 66.29 & 33.33 & 65.67 & 96.53 & 73.33 & 15.89 \\
Qwen3-14B-Thinking & 69.00 & 93.50 & 66.67 & 90.58 & 100.00 & 100.00 & 82.67 & 73.04 & 89.46 & 66.67 & 85.57 & 100.00 & 100.00 & 71.89 & 84.75 & 77.58 & 58.33 & 73.65 & 99.87 & 99.45 & 40.44 & 85.67 & 76.83 & 66.67 & 72.96 & 100.00 & 100.00 & 38.22 \\
Qwen3-32B & 94.58 & 67.42 & 58.33 & 65.81 & 99.60 & 95.38 & 13.78 & 96.75 & 65.42 & 75.00 & 64.43 & 99.73 & 94.87 & 8.22 & 94.00 & 66.00 & 58.33 & 65.16 & 98.00 & 79.17 & 12.67 & 96.58 & 65.50 & 79.17 & 64.50 & 99.67 & 93.90 & 8.56 \\
Qwen3-32B-Thinking & 87.71 & 74.79 & 62.50 & 71.26 & 100.00 & 100.00 & 32.78 & 90.79 & 71.71 & 62.50 & 68.84 & 100.00 & 100.00 & 24.56 & 87.67 & 74.83 & 62.50 & 71.29 & 100.00 & 100.00 & 32.89 & 88.83 & 73.58 & 58.33 & 70.31 & 99.93 & 99.63 & 29.67 \\

Qwen3-30B-A3B-Instruct & 71.04 & 91.04 & 62.50 & 87.68 & 99.67 & 99.28 & 76.67 & 77.71 & 83.62 & 62.50 & 79.68 & 99.07 & 97.38 & 57.89 & 75.12 & 86.62 & 58.33 & 82.70 & 99.40 & 98.49 & 65.33 & 84.50 & 77.75 & 58.33 & 73.82 & 99.80 & 99.19 & 41.00 \\
Qwen3-30B-A3B-Thinking & 84.12 & 78.38 & 62.50 & 74.29 & 100.00 & 100.00 & 42.33 & 90.75 & 71.75 & 62.50 & 68.87 & 100.00 & 100.00 & 24.67 & 85.17 & 77.33 & 62.50 & 73.39 & 100.00 & 100.00 & 39.56 & 84.92 & 77.58 & 62.50 & 73.60 & 100.00 & 100.00 & 40.22 \\
Qwen3-NEXT-80B-A3B-instruct & 73.50 & 88.92 & 62.50 & 84.98 & 99.93 & 99.84 & 70.56 & 74.71 & 87.79 & 66.67 & 83.66 & 100.00 & 100.00 & 67.44 & 65.08 & 97.42 & 66.67 & 96.03 & 100.00 & 100.00 & 93.11 & 66.04 & 96.46 & 66.67 & 94.64 & 100.00 & 100.00 & 90.56 \\
Qwen3-NEXT-80B-A3B-Thinking & 74.83 & 87.67 & 66.67 & 83.52 & 100.00 & 100.00 & 67.11 & 76.54 & 85.96 & 66.67 & 81.65 & 100.00 & 100.00 & 62.56 & 69.79 & 92.54 & 62.50 & 89.55 & 100.00 & 100.00 & 80.47 & 68.58 & 93.54 & 70.83 & 91.13 & 100.00 & 100.00 & 83.61 \\
Qwen3-235B-A22B-Instruct & 80.96 & 81.21 & 54.17 & 76.99 & 99.73 & 99.12 & 50.33 & 83.67 & 78.50 & 54.17 & 74.50 & 99.73 & 98.98 & 43.11 & 73.21 & 89.21 & 58.33 & 85.32 & 99.93 & 99.84 & 71.33 & 78.46 & 84.04 & 66.67 & 79.66 & 100.00 & 100.00 & 57.44 \\
Qwen3-235B-A22B-Thinking & 69.50 & 93.00 & 66.67 & 89.93 & 100.00 & 100.00 & 81.33 & 72.00 & 90.50 & 66.67 & 86.81 & 100.00 & 100.00 & 74.67 & 71.38 & 91.12 & 66.67 & 87.57 & 100.00 & 100.00 & 76.33 & 73.04 & 89.46 & 62.50 & 85.57 & 100.00 & 100.00 & 71.89 \\

Gemma-3-1B-IT & 75.38 & 51.04 & 0.00 & 58.98 & 71.13 & 26.73 & 17.56 & 82.29 & 55.79 & 0.00 & 61.11 & 80.47 & 31.06 & 14.67 & 78.92 & 45.92 & 0.00 & 55.33 & 69.87 & 10.67 & 6.00 & 83.71 & 50.79 & 0.00 & 57.94 & 77.60 & 14.07 & 6.11 \\
Gemma-3-4B-IT & 97.83 & 62.25 & 33.33 & 62.65 & 98.07 & 44.23 & 2.56 & 97.46 & 61.71 & 33.33 & 62.42 & 97.33 & 34.43 & 2.33 & 95.33 & 63.83 & 8.33 & 63.81 & 97.33 & 64.29 & 8.00 & 97.92 & 64.17 & 50.00 & 63.62 & 99.67 & 90.00 & 5.00 \\
Gemma-3-12B-IT & 99.21 & 62.71 & 75.00 & 62.70 & 99.53 & 63.16 & 1.33 & 98.71 & 63.54 & 70.83 & 63.19 & 99.80 & 90.32 & 3.11 & 99.75 & 62.58 & 83.33 & 62.57 & 99.87 & 66.67 & 0.44 & 100.00 & 62.50 & 100.00 & 62.50 & 100.00 & 0.00 & 0.00 \\

Gemma-3-27B-IT & 97.62 & 64.29 & 54.17 & 63.72 & 99.53 & 87.72 & 5.56 & 98.04 & 63.71 & 37.50 & 63.37 & 99.40 & 80.85 & 4.22 & 99.08 & 63.25 & 79.17 & 62.99 & 99.87 & 90.91 & 2.22 & 99.92 & 62.58 & 95.83 & 62.55 & 100.00 & 100.00 & 0.22 \\
Llama3-8B-Instruct & 59.25 & 62.71 & 0.00 & 71.31 & 67.60 & 50.26 & 54.62 & 64.33 & 59.29 & 0.00 & 66.97 & 68.93 & 45.50 & 43.27 & 35.96 & 53.08 & 0.00 & 71.73 & 41.29 & 42.67 & 72.86 & 42.42 & 51.62 & 0.00 & 66.70 & 45.30 & 40.58 & 62.29 \\

Llama3.3-70B-Instruct & 98.12 & 64.29 & 87.50 & 63.65 & 99.93 & 97.78 & 4.89 & 98.38 & 63.21 & 58.33 & 63.07 & 99.27 & 71.79 & 3.11 & 98.58 & 63.33 & 87.50 & 63.10 & 99.53 & 79.41 & 3.00 & 99.62 & 62.46 & 79.17 & 62.53 & 99.67 & 44.44 & 0.44 \\
Llama3-70B-Instruct & 99.12 & 63.12 & 83.33 & 62.93 & 99.80 & 85.71 & 2.00 & 98.21 & 62.62 & 66.67 & 62.79 & 98.67 & 53.49 & 2.56 & 94.96 & 63.54 & 41.67 & 63.71 & 96.80 & 60.33 & 8.11 & 90.83 & 61.25 & 20.83 & 63.07 & 91.67 & 43.18 & 10.56 \\

\bottomrule
\end{tabular}
}
\caption{The detailed results of baseline prompt of open-source models.}
\label{tab:by_lang_concept_cg_detailed}
\end{table*}

\subsubsection{The external experiment of thinking}
\label{thinking_external}
The Instruct+ CoT experiment is shown in Tab.\ref{tab:Instruct_CoT}.
The \texttt{Instruct+CoT} setting can induce a partial shift toward modern logic, but the shift is limited.
The experiments of DeepSeek-R1 and DeepSeek-R1-Distill models are shown in Table~\ref{tab:ds_ds_distill}. RL training does not automatically lead to rigorous modern logic in all models.

\begin{table*}[t]
\centering
\resizebox{\textwidth}{!}{%
\begin{tabular}{l|ccccccc|ccccccc|ccccccc|ccccccc}
\toprule
Model & \multicolumn{7}{c}{ZH+} & \multicolumn{7}{c}{ZH-} & \multicolumn{7}{c}{EN+} & \multicolumn{7}{c}{EN-} \\
\cmidrule(lr){2-8}
\cmidrule(lr){9-15}
\cmidrule(lr){16-22}
\cmidrule(lr){23-29}
  & $\text{Acc}_{t}$ & $\text{Acc}_{m}$ & Cons & $\mathrm{pre}_V$ & $\mathrm{rec}_V$ & $\mathrm{pre}_I$ & $\mathrm{rec}_I$ & $\text{Acc}_{t}$ & $\text{Acc}_{m}$ & Cons & $\mathrm{pre}_V$ & $\mathrm{rec}_V$ & $\mathrm{pre}_I$ & $\mathrm{rec}_I$ & $\text{Acc}_{t}$ & $\text{Acc}_{m}$ & Cons & $\mathrm{pre}_V$ & $\mathrm{rec}_V$ & $\mathrm{pre}_I$ & $\mathrm{rec}_I$ & $\text{Acc}_{t}$ & $\text{Acc}_{m}$ & Cons & $\mathrm{pre}_V$ & $\mathrm{rec}_V$ & $\mathrm{pre}_I$ & $\mathrm{rec}_I$ \\
\midrule
Qwen3-0.6B & 98.75 & 62.25 & 50.00 & 62.53 & 98.80 & 40.00 & 1.33 & 97.38 & 61.96 & 16.67 & 62.56 & 97.47 & 39.68 & 2.78 & 99.96 & 62.46 & 95.83 & 62.48 & 99.93 & 0.00 & 0.00 & 99.21 & 62.62 & 37.50 & 62.66 & 99.47 & 57.89 & 1.22 \\
Qwen3-1.7B & 93.50 & 64.17 & 12.50 & 64.26 & 96.13 & 62.82 & 10.89 & 92.00 & 62.50 & 16.67 & 63.59 & 93.60 & 50.00 & 10.67 & 67.08 & 56.50 & 0.00 & 64.16 & 68.87 & 40.89 & 35.89 & 58.88 & 51.71 & 0.00 & 62.07 & 58.47 & 36.88 & 40.44 \\
Qwen3-4B & 91.12 & 66.88 & 25.00 & 66.12 & 96.40 & 74.65 & 17.67 & 94.54 & 65.71 & 33.33 & 64.92 & 98.20 & 79.39 & 11.56 & 55.67 & 54.42 & 0.00 & 65.19 & 58.11 & 40.92 & 48.33 & 72.75 & 53.08 & 0.00 & 60.71 & 70.67 & 32.72 & 23.78 \\
Qwen3-8B & 88.83 & 70.75 & 25.00 & 68.71 & 97.67 & 86.94 & 25.89 & 94.54 & 67.12 & 50.00 & 65.67 & 99.33 & 92.37 & 13.44 & 86.08 & 72.58 & 4.17 & 70.38 & 96.93 & 86.23 & 32.00 & 90.33 & 66.50 & 4.17 & 66.05 & 95.47 & 70.69 & 18.22 \\
Qwen3-14B & 88.08 & 74.00 & 54.17 & 70.72 & 99.67 & 98.25 & 31.22 & 94.42 & 67.67 & 54.17 & 65.98 & 99.67 & 96.27 & 14.33 & 85.42 & 72.08 & 16.67 & 70.24 & 96.00 & 82.86 & 32.22 & 90.58 & 70.42 & 33.33 & 68.17 & 98.80 & 92.04 & 23.11 \\
Qwen3-32B & 89.58 & 72.50 & 50.00 & 69.53 & 99.67 & 98.00 & 27.22 & 94.17 & 68.33 & 75.00 & 66.37 & 100.00 & 100.00 & 15.56 & 89.12 & 72.21 & 41.67 & 69.47 & 99.07 & 94.64 & 27.44 & 91.62 & 70.12 & 37.50 & 67.80 & 99.40 & 95.52 & 21.33 \\

Qwen3-30B-A3B-Instruct & 63.88 & 98.62 & 75.00 & 97.85 & 100.00 & 100.00 & 96.33 & 67.25 & 95.25 & 62.50 & 92.94 & 100.00 & 100.00 & 87.33 & 63.88 & 98.62 & 75.00 & 97.85 & 100.00 & 100.00 & 96.33 & 66.54 & 95.96 & 66.67 & 93.93 & 100.00 & 100.00 & 89.22 \\
Qwen3-Next-80B-A3B-Instruct & 65.46 & 97.04 & 66.67 & 95.48 & 100.00 & 100.00 & 92.11 & 71.33 & 91.17 & 62.50 & 87.62 & 100.00 & 100.00 & 76.44 & 62.75 & 99.75 & 83.33 & 99.60 & 100.00 & 100.00 & 99.33 & 62.79 & 99.62 & 79.17 & 99.47 & 100.00 & 100.00 & 99.11 \\
Qwen3-235B-A22B-Instruct & 67.38 & 95.12 & 70.83 & 92.76 & 100.00 & 100.00 & 87.00 & 71.29 & 91.21 & 62.50 & 87.67 & 100.00 & 100.00 & 76.56 & 62.62 & 99.88 & 87.50 & 99.80 & 100.00 & 100.00 & 99.67 & 62.75 & 99.75 & 83.33 & 99.60 & 100.00 & 100.00 & 99.33 \\

\bottomrule
\end{tabular}
}
\caption{Instruct+CoT setting experiment.}
\label{tab:Instruct_CoT}
\end{table*}


\begin{table*}[t]
\centering
\resizebox{\textwidth}{!}{%
\begin{tabular}{l|ccc|ccc|ccc|ccc}
\toprule
Model & \multicolumn{3}{c}{ZH+} & \multicolumn{3}{c}{ZH-} & \multicolumn{3}{c}{EN+} & \multicolumn{3}{c}{EN-} \\
\cmidrule(lr){2-4}
\cmidrule(lr){5-7}
\cmidrule(lr){8-10}
\cmidrule(lr){11-13}
  & $\text{Acc}_{t}$ & $\text{Acc}_{m}$ & Cons & $\text{Acc}_{t}$ & $\text{Acc}_{m}$ & Cons & $\text{Acc}_{t}$ & $\text{Acc}_{m}$ & Cons & $\text{Acc}_{t}$ & $\text{Acc}_{m}$ & Cons \\
\midrule

DeepSeek-R1 & 76.00 & 86.50 & 62.50 & 78.83 & 83.67 & 62.50 & 73.96 & 88.50 & 62.50 & 77.54 & 84.88 & 58.33 \\
DeepSeek-R1-Distill-Llama-8B & 99.00 & 62.79 & 54.17 & 99.04 & 62.79 & 45.83 & 94.62 & 61.71 & 4.17 & 96.21 & 61.46 & 12.50 \\
DeepSeek-R1-Distill-Llama-70B & 96.75 & 65.42 & 58.33 & 98.12 & 64.12 & 62.50 & 95.42 & 65.88 & 29.17 & 97.42 & 64.25 & 45.83 \\
DeepSeek-R1-Distill-Qwen-14B & 99.54 & 62.88 & 83.33 & 99.67 & 62.42 & 79.17 & 99.42 & 62.54 & 58.33 & 99.54 & 62.54 & 70.83 \\
\bottomrule
\end{tabular}
}
\caption{The DeepSeek-R1 and DeepSeek-R1-Distilled results by language and concept existence.}
\label{tab:ds_ds_distill}
\end{table*}

\subsubsection{The results of Base models}
\label{base_model}
The experiments of Base models are shown in Table~\ref{tab:base_model_results_detailed}. Base models are the starting point and a constraint to further training.

\begin{table*}[t]
\centering
\resizebox{\textwidth}{!}{%
\begin{tabular}{lcccccccccccc}
\toprule
Model & \multicolumn{3}{c}{ZH+} & \multicolumn{3}{c}{ZH-} & \multicolumn{3}{c}{EN+} & \multicolumn{3}{c}{EN-} \\
\cmidrule(lr){2-4}
\cmidrule(lr){5-7}
\cmidrule(lr){8-10}
\cmidrule(lr){11-13}
  & $\text{Acc}_{t}$ & $\text{Acc}_{m}$ & $\text{Cons}$ & $\text{Acc}_{t}$ & $\text{Acc}_{m}$ & \text{Cons} & $\text{Acc}_{t}$ & $\text{Acc}_{m}$ & \text{Cons} & $\text{Acc}_{t}$ & $\text{Acc}_{m}$ & \text{Cons} \\
\midrule
Qwen3-0.6B-Base & 6.62 & 36.46 & 29.17 & 14.67 & 34.00 & 4.17 & 1.71 & 37.38 & 16.67 & 7.42 & 36.92 & 0.00 \\
Qwen3-1.7B-Base & 82.88 & 55.79 & 0.00 & 77.21 & 52.54 & 0.00 & 99.92 & 62.58 & 91.67 & 99.96 & 62.54 & 95.83 \\
Qwen3-4B-Base & 93.25 & 61.67 & 33.33 & 91.29 & 60.54 & 29.17 & 72.54 & 56.54 & 8.33 & 81.58 & 58.83 & 25.00 \\
Qwen3-8B-Base & 95.00 & 61.17 & 16.67 & 81.42 & 53.67 & 0.00 & 73.75 & 68.21 & 33.33 & 82.92 & 64.33 & 0.00 \\
Qwen3-30B-A3B-Base & 79.50 & 52.12 & 0.00 & 90.17 & 58.00 & 0.00 & 79.58 & 50.75 & 0.00 & 84.96 & 53.67 & 0.00 \\

Gemma-3-1B-PT & 6.63 & 32.53 & 0.00 & 5.38 & 24.82 & 0.00 & 5.12 & 21.90 & 0.00 & 5.11 & 23.37 & 0.00 \\
Gemma-3-4B-PT & 6.75 & 36.67 & 0.00 & 9.12 & 36.08 & 0.00 & 2.42 & 30.31 & 0.00 & 3.48 & 30.74 & 0.00 \\
Gemma-3-12B-PT & 9.00 & 39.00 & 0.00 & 11.33 & 38.83 & 0.00 & 11.83 & 38.79 & 0.00 & 14.49 & 38.44 & 0.00 \\
Gemma-3-27B-PT & 26.83 & 42.50 & 0.00 & 24.46 & 42.29 & 0.00 & 16.75 & 34.21 & 0.00 & 16.92 & 35.42 & 0.00 \\

Llama3-8B-Base & 30.08 & 36.67 & 0.00 & 29.58 & 35.54 & 0.00 & 12.17 & 32.92 & 0.00 & 13.54 & 34.42 & 0.00 \\
Llama3-70B-Base & 44.70 & 43.44 & 0.00 & 41.79 & 42.71 & 0.00 & 34.43 & 45.78 & 0.00 & 29.86 & 43.65 & 0.00 \\
\bottomrule
\end{tabular}
}
\caption{The results of various Base Models by language and concept existence.}
\label{tab:base_model_results_detailed}
\end{table*}

\subsubsection{The results of dLLMs}

We conduct experiments on various dLLMs, including LLaDA \cite{nie2025largelanguagediffusionmodels}, LLaDA-1.5 \cite{zhu2025llada15variancereducedpreference}, TraDo \cite{wang2025revolutionizing}, DiRL \cite{zhu2025dirl}, SDAR \cite{cheng2025sdar}, LLaDA2.0 \cite{bie2025llada20}. The results are shown in Table\ref{tab:dllm_by_lang_concept}.

\begin{table*}[ht]
\centering
\resizebox{\textwidth}{!}{%
\begin{tabular}{lcccccccccccc}
\toprule
Model & \multicolumn{3}{c}{ZH+} & \multicolumn{3}{c}{ZH-} & \multicolumn{3}{c}{EN+} & \multicolumn{3}{c}{EN-} \\
\cmidrule(lr){2-4}
\cmidrule(lr){5-7}
\cmidrule(lr){8-10}
\cmidrule(lr){11-13}
  & $\text{Acc}_{t}$ & $\text{Acc}_{m}$ & \text{Cons} & $\text{Acc}_{t}$ & $\text{Acc}_{m}$ & \text{Cons} & $\text{Acc}_{t}$ & $\text{Acc}_{m}$ & \text{Cons} & $\text{Acc}_{t}$ & $\text{Acc}_{m}$ & \text{Cons} \\
\midrule
LLaDA-8b-Instruct & 70.25 & 50.83 & 0.00 & 68.79 & 50.54 & 0.00 & 71.75 & 55.88 & 0.00 & 69.50 & 54.12 & 0.00 \\
LLaDA-1.5 & 73.12 & 50.62 & 0.00 & 75.08 & 52.88 & 0.00 & 66.46 & 57.79 & 0.00 & 61.25 & 53.25 & 0.00 \\
TraDo-4B-Instruct & 84.92 & 59.50 & 20.83 & 80.67 & 57.00 & 25.00 & 87.04 & 66.62 & 20.83 & 86.29 & 63.96 & 12.50 \\
TraDo-8B-Instruct & 96.38 & 61.04 & 41.67 & 95.04 & 59.71 & 20.83 & 90.92 & 69.50 & 37.50 & 89.46 & 66.88 & 4.17 \\
DiRL-8B-Instruct & 89.50 & 62.19 & 0.00 & 92.83 & 59.00 & 0.00 & 94.12 & 66.25 & 20.83 & 94.58 & 63.25 & 4.17 \\
SDAR-4B & 80.46 & 59.04 & 20.83 & 76.08 & 56.50 & 16.67 & 78.33 & 63.67 & 16.67 & 76.96 & 62.21 & 12.50 \\
SDAR-8B & 91.58 & 59.17 & 20.83 & 90.83 & 56.33 & 20.83 & 84.21 & 68.46 & 0.00 & 72.75 & 62.42 & 0.00 \\
SDAR-30B-A3B & 99.17 & 63.17 & 79.17 & 99.00 & 63.33 & 79.17 & 99.71 & 62.71 & 75.00 & 99.50 & 62.83 & 70.83 \\
LLaDA2.0-mini & 82.62 & 77.46 & 16.67 & 87.12 & 74.46 & 33.33 & 85.96 & 73.96 & 37.50 & 89.42 & 72.08 & 45.83 \\
LLaDA2.0-flash & 73.21 & 89.17 & 62.50 & 80.46 & 81.96 & 58.33 & 72.04 & 90.46 & 66.67 & 76.54 & 85.88 & 62.50 \\
\bottomrule
\end{tabular}
}
\caption{The brief results of various dLLMs by language and concept existence.}
\label{tab:dllm_by_lang_concept}
\end{table*}

\begin{table*}[t]
\centering
\resizebox{\textwidth}{!}{%
\begin{tabular}{lcccccccccccccccccccccccccccc}
\toprule
Model & \multicolumn{7}{c}{ZH+} & \multicolumn{7}{c}{ZH-} & \multicolumn{7}{c}{EN+} & \multicolumn{7}{c}{EN-} \\
\cmidrule(lr){2-8}
\cmidrule(lr){9-15}
\cmidrule(lr){16-22}
\cmidrule(lr){23-29}
  & $\text{Acc}_{t}$ & $\text{Acc}_{m}$ & Cons & $\mathrm{pre}_V$ & $\mathrm{rec}_V$ & $\mathrm{pre}_I$ & $\mathrm{rec}_I$
  & $\text{Acc}_{t}$ & $\text{Acc}_{m}$ & Cons & $\mathrm{pre}_V$ & $\mathrm{rec}_V$ & $\mathrm{pre}_I$ & $\mathrm{rec}_I$
  & $\text{Acc}_{t}$ & $\text{Acc}_{m}$ & Cons & $\mathrm{pre}_V$ & $\mathrm{rec}_V$ & $\mathrm{pre}_I$ & $\mathrm{rec}_I$
  & $\text{Acc}_{t}$ & $\text{Acc}_{m}$ & Cons & $\mathrm{pre}_V$ & $\mathrm{rec}_V$ & $\mathrm{pre}_I$ & $\mathrm{rec}_I$ \\
\midrule
Qwen3-0.6B-Base & 6.62 & 36.46 & 29.17 & 42.14 & 4.47 & 36.06 & 89.78 & 14.67 & 34.00 & 4.17 & 38.92 & 9.20 & 33.43 & 75.95 & 1.71 & 37.38 & 16.67 & 51.22 & 1.40 & 37.21 & 97.77 & 7.42 & 36.92 & 0.00 & 50.00 & 5.99 & 36.33 & 89.95 \\
Qwen3-1.7B-Base & 82.88 & 55.79 & 0.00 & 62.95 & 90.40 & 39.55 & 10.56 & 77.21 & 52.54 & 0.00 & 62.28 & 87.29 & 38.91 & 13.28 & 99.92 & 62.58 & 91.67 & 62.55 & 100.00 & 100.00 & 0.22 & 99.96 & 62.54 & 95.83 & 62.53 & 100.00 & 100.00 & 0.11 \\
Qwen3-4B-Base & 93.25 & 61.67 & 33.33 & 62.96 & 93.93 & 43.83 & 7.89 & 91.29 & 60.54 & 29.17 & 62.62 & 91.47 & 38.76 & 9.00 & 72.54 & 56.54 & 8.33 & 63.12 & 73.27 & 39.15 & 28.67 & 81.58 & 58.83 & 25.00 & 63.07 & 82.33 & 40.05 & 19.67 \\
Qwen3-8B-Base & 95.00 & 61.17 & 16.67 & 62.54 & 95.58 & 38.89 & 4.69 & 81.42 & 53.67 & 0.00 & 59.93 & 78.28 & 26.47 & 13.00 & 73.75 & 68.21 & 33.33 & 70.85 & 83.60 & 60.89 & 42.60 & 82.92 & 64.33 & 0.00 & 66.18 & 87.92 & 55.64 & 25.22 \\
Qwen3-30B-A3B-Base & 79.50 & 52.12 & 0.00 & 63.31 & 96.33 & 48.31 & 5.79 & 90.17 & 58.00 & 0.00 & 62.62 & 96.99 & 46.84 & 4.37 & 79.58 & 50.75 & 0.00 & 62.51 & 98.60 & 58.54 & 3.24 & 84.96 & 53.67 & 0.00 & 62.53 & 99.07 & 52.00 & 1.67 \\

Gemma-3-1B-PT & 6.63 & 32.53 & 0.00 & 63.52 & 8.27 & 37.74 & 92.13 & 5.38 & 24.82 & 0.00 & 59.69 & 7.82 & 36.33 & 90.88 & 5.12 & 21.90 & 0.00 & 65.57 & 9.83 & 37.59 & 91.32 & 5.11 & 23.37 & 0.00 & 65.57 & 9.15 & 37.58 & 91.92 \\
Gemma-3-4B-PT & 6.75 & 36.67 & 0.00 & 65.43 & 7.61 & 37.55 & 93.25 & 9.12 & 36.08 & 0.00 & 60.73 & 9.61 & 36.95 & 89.50 & 2.42 & 30.31 & 0.00 & 52.63 & 2.62 & 38.01 & 96.20 & 3.48 & 30.74 & 0.00 & 60.49 & 4.12 & 36.93 & 95.42 \\
Gemma-3-12B-PT & 9.00 & 39.00 & 0.00 & 65.74 & 9.71 & 37.56 & 91.47 & 11.33 & 38.83 & 0.00 & 61.76 & 11.61 & 37.40 & 88.02 & 11.83 & 38.79 & 0.00 & 64.79 & 13.18 & 38.13 & 88.19 & 14.49 & 38.44 & 0.00 & 63.87 & 15.79 & 37.15 & 84.79 \\
Gemma-3-27B-PT & 26.83 & 42.50 & 0.00 & 62.11 & 27.99 & 37.60 & 71.76 & 24.46 & 42.29 & 0.00 & 63.37 & 26.01 & 37.80 & 74.94 & 16.75 & 34.21 & 0.00 & 62.19 & 20.59 & 37.20 & 78.98 & 16.92 & 35.42 & 0.00 & 65.76 & 22.05 & 38.18 & 80.75 \\

Llama3-8B-Base & 30.08 & 36.67 & 0.00 & 63.99 & 40.46 & 38.07 & 61.65 & 29.58 & 35.54 & 0.00 & 60.28 & 38.01 & 37.85 & 60.11 & 12.17 & 32.92 & 0.00 & 60.96 & 14.86 & 37.50 & 84.30 & 13.54 & 34.42 & 0.00 & 63.69 & 17.05 & 38.07 & 83.99 \\
Llama3-70B-Base & 44.70 & 43.44 & 0.00 & 62.85 & 53.47 & 38.57 & 48.03 & 41.79 & 42.71 & 0.00 & 62.70 & 48.23 & 36.99 & 51.43 & 34.43 & 45.78 & 0.00 & 64.71 & 37.99 & 39.26 & 65.92 & 29.86 & 43.65 & 0.00 & 64.62 & 32.99 & 38.34 & 69.76 \\

\bottomrule
\end{tabular}
}
\caption{The detailed results of Base models.}
\label{tab:by_lang_concept_base_detailed}
\end{table*}

\end{document}